\documentclass{article}

\PassOptionsToPackage{numbers}{natbib}
\usepackage[preprint]{neurips_2022}

\usepackage[utf8]{inputenc} % allow utf-8 input
\usepackage[T1]{fontenc}    % use 8-bit T1 fonts
\usepackage{hyperref}       % hyperlinks
\usepackage{url}            % simple URL typesetting
\usepackage{booktabs}       % professional-quality tables
\usepackage{amsfonts}       % blackboard math symbols
\usepackage{nicefrac}       % compact symbols for 1/2, etc.
\usepackage{microtype}      % microtypography
\usepackage{xcolor}         % colors
\usepackage{booktabs}

\usepackage{array}
\usepackage{microtype}
\usepackage{graphicx}
\usepackage{bm}
\usepackage{algorithm}
\usepackage{algorithmic}
\usepackage{multirow}
\usepackage{amsthm}
\usepackage{amsmath}
\usepackage{subfigure}
\usepackage{amssymb}
\usepackage{bbding}
\usepackage{pifont}
\usepackage{wasysym}
\usepackage{booktabs} % for professional tables
\usepackage{makecell}
\usepackage{subfigure}
\usepackage{bbding}
\usepackage{balance}
\usepackage{float}
\usepackage{subeqnarray}
\usepackage{cases}
\usepackage{tabularx}
\usepackage{color}
\usepackage{colortbl,url}
\usepackage{wrapfig}
\usepackage{threeparttable}
\newcommand{\cmark}{\ding{51}}%
\newcommand{\xmark}{\ding{55}}%

\title{Revisiting GANs by Best-Response Constraint: Perspective, Methodology, and Application}

\author{%
	Risheng~Liu$^{\ast 1,2}\quad $ Jiaxin~Gao$^{1}\quad$   Xuan~Liu$^{1}\quad$  Xin~Fan$^{1,2}$\\
	$^{1}$International School of Information Science \& Engineering, DUT\\
}

\begin{document}

\maketitle

\begin{abstract}
	%Recent research on adversarial learning mechanisms,  specifically Generative Adversarial Networks (GANs), has achieved great success in vision and learning areas. The vanilla GAN implements a divergence minimization training procedure from an adversarial perspective by introducing extra discriminators. Considering the single-level minimax objective and alternating optimization strategies often tend to cause mode collapse, vanishing gradients, and oscillations in the training phase, we propose a novel hierarchical learning perspective to reveal the operating mechanism of GAN. Specifically, we define a Best-Response Constraint (BRC) to accurately portray the parameter coupling relationship between the generator and the discriminator, forcing an accurately estimated optimal discriminator back to the generator. On this basis, a fast implicit gradient approximation strategy that combines an outer-product-based Hessian approximation technique is constructed to break away from the shackles of the inherent training mechanism.  Noteworthy, our method can serve as a plug-and-play drop-in replacement that can be flexibly embedded into most of the advanced GAN variants, profiting from the fault tolerance of this novel optimization paradigm for objective function requirements. Quantitative and qualitative synthetic and real-world experimental results demonstrate that our approach maintains the characteristics of low computational complexity, flexible applicability, stable and fast convergence. 
	In past years, the minimax type single-level optimization formulation and its variations have been widely utilized to address Generative Adversarial Networks (GANs). Unfortunately, it has been proved that these alternating learning strategies cannot exactly reveal the intrinsic relationship between the generator and discriminator, thus easily result in a series of issues, including mode collapse, vanishing gradients and oscillations in the training phase, etc. In this work, by investigating the fundamental mechanism of GANs from the perspective of hierarchical optimization, we propose Best-Response Constraint (BRC), a general learning framework, that can explicitly formulate the potential dependency of the generator on the discriminator. Rather than adopting these existing time-consuming bilevel iterations, we design an implicit gradient scheme with outer-product Hessian approximation as our fast solution strategy.  \emph{Noteworthy, we demonstrate that even with different motivations and formulations, a variety of existing GANs ALL can be uniformly improved by our flexible BRC methodology.} Extensive quantitative and qualitative experimental results verify the effectiveness, flexibility and stability of our proposed framework.

\end{abstract}

\section{Introduction} \label{intro}
Generative Adversarial Networks (GANs), one of the most popular frameworks of generative models, aim to construct the problem as a adversarial learning procedure with the generator and the discriminator. 
Under the adversarial learning mechanism, Vanilla GAN (VGAN) training procedure seeks to find the solution to a minimax formulation by means of an alternating gradient strategy greedily~\cite{goodfellow2014generative}.
In recent years, GAN and its variants have been exploited with significant success for a broad spectrum of complex vision and learning problems, such as image generation~\cite{li2019controllable,zhu2017unpaired}, low-level image and video enhancement~\cite{ledig2017photo,wang2018esrgan}, high-level recognition and  detection~\cite{marriott20213d,tran2017disentangled}, and reinforcement learning~\cite{pfau2016connecting,chen2020computation}, just to name a few. 
Formally, the goal of GANs is to produce a generator network $G$ (with parameters $\bm{\theta}_G$) that is able to accurately reveal some unknown data distribution $P_{data}$. It can be typically analyzed from the divergence minimization perspective 
\begin{equation}
	\min_{\bm{\theta}_{G}}``\mathtt{distance}"(P_{G},P_{data})\rightarrow \min_{\bm{\theta}_{G}}\max_{\bm{\theta}_{D}}F(\bm{\theta}_G,\bm{\theta}_D).\label{eq:distance}
\end{equation}
Here, $``\mathtt{distance}"$ generally denotes some given statistical distance (e.g., Jensen-Shannon divergence, f-divergence, and Wasserstein distance) between the distribution of the generated data $P_{G}$ and the latent distribution $P_{data}$. 
By introducing an adversarial discriminator $D$ (parameter $\bm{\theta}_D$), VGAN dynamics defines a single-level minimax multivariate function that includes the competition of the two networks.  
A further explanation is to utilize the above-mentioned maximum objective on the discriminator to assist in computing the divergence minimization process. However, this modeling formulation of GANs puts the discriminator $D$ and the generator $G$ in the identical status, and could not characterize the coupling relationship between the generator $G$ and the discriminator $D$, thus is always notoriously fragile and unstable. Following the widespread adoption of joint cross-entropy loss in VGAN and its variants, i.e., $
F(\bm{\theta}_G,\bm{\theta}_D)=E_{\mathbf{x}\sim P_{data}}\left[\log(D(\mathbf{x}))\right] 
+ E_{\mathbf{z}\sim P_{\mathbf{z}}}\left[\log(1-D(G(\mathbf{z})))\right], 
$ 
various variations of objective function $F$ suitable for specific applications have emerged.

% to relieve the discrepancy in training process
In recent years, researchers are keen on a variety of GAN improvements mainly including four categories, selecting suitable loss functions~\cite{arjovsky2017wasserstein,mao2017least}, utilizing better normalization~\cite{miyato2018spectral,che2016mode}, designing delicate architectures~\cite{heusel2017gans,zhu2017unpaired} and implementing various training strategies~\cite{metz2016unrolled,farnia2020gans}. 
%a plethora of extensions of GAN variants have been established with different formulations and strategies, for instance, selecting suitable loss functions, utilizing better normalization, and implementing various training iteration procedures. 
Some approaches extended in the context of VGAN, combine spectral normalization with gradient penalty to prevent the discriminator from learning too fast and triggering gradient disappearance~\cite{miyato2018spectral,arjovsky2017wasserstein}. Other methods~\cite{salimans2016improved,tolstikhin2017adagan} introduce regularization of the gradient norm to improve the diversity of the generated samples, thus improving the robustness of the framework under different structures.  
%In addition, the choice of hyper-parameters, such as batch size, momentum, weight decay and learning rate, are also the most important factors for stable training~\cite{salimans2016improved}. 
Although numerous GAN variants coming from different disciplines and perspectives exist in the previous literature, it still remains unclear how these techniques combine with each other and how the combination affects the overall performance. Conversely,  significant manual parameter tuning and redundant network structures pose a significant obstacle to the wider adoption of current GANs as a off-the-shelf framework.  
Intuitively speaking, current common failures regarding training GANs include generating poor generation quality, vanishing or exploding gradients, training instability, and mode collapse~\cite{pavan2021generative}. To date none of these methods are completely effective in alleviating all the above difficulties. 
\begin{figure*}[htbp]
	%\centering 
	\begin{tabular}{l}
		\includegraphics[width=0.99\linewidth]{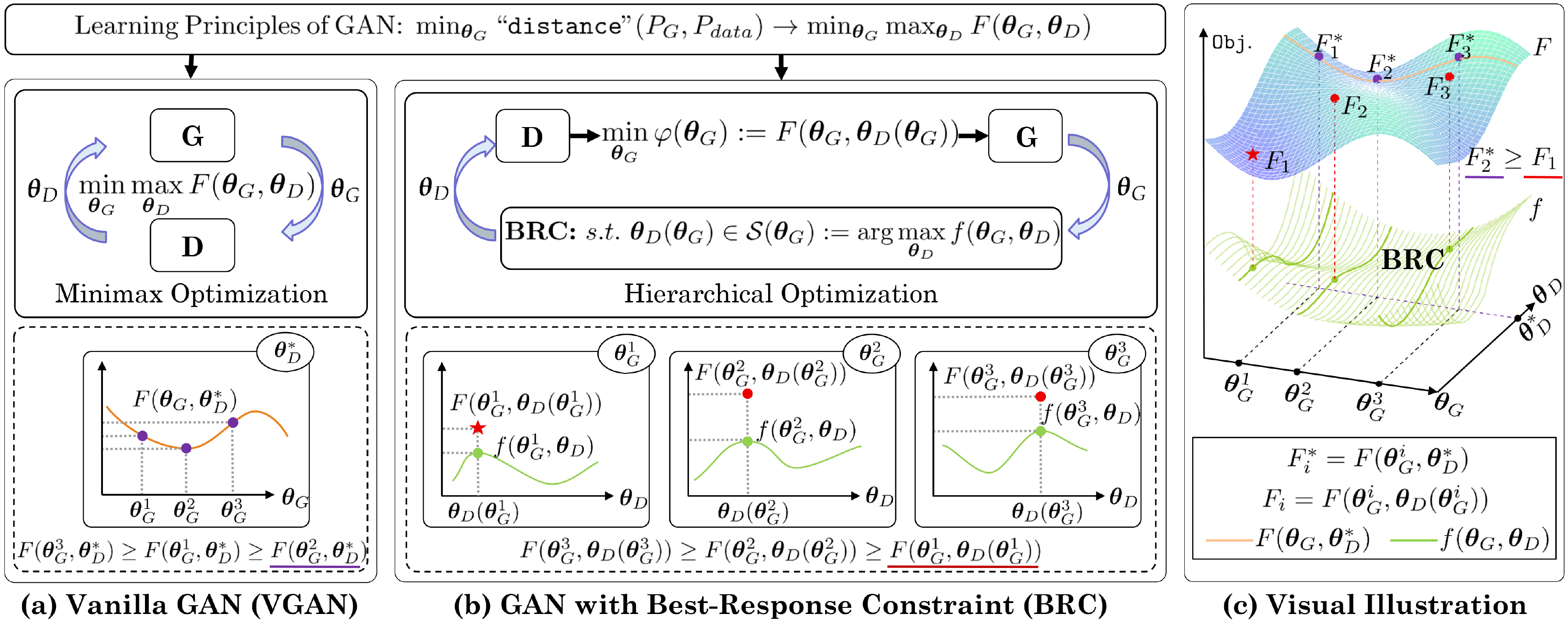} 
\end{tabular}%
\caption{Illustration of the learning principle of GAN. The traditional VGAN (a) and our proposed BRC modeling framework (b), as well as the solution strategy (c) are displayed from left to right. The dashed boxes exhibit why our proposed BRC framework can find the optimal solution from three generator variables compared to VGAN. 
	%We provide an example demonstrating how to find the optimal solution from three generator variables by executing two methods separately, as indicated by the dashed box.
	%Comparisons of modelling and solving strategy under different optimization paradigms (i.e., VGAN and GAN with BRC). By sampling three generator parameters (i.e., $\theta_{G}^1, \theta_{G}^2, \theta_{G}^3$), we provide an example demonstrating how to find the optimal solution from three variables by executing two methods separately, as indicated by the dashed box. Even for a locally optimal discriminator parameter $\theta_{D}^{*}$, VGAN may only find a suboptimal result $\theta_{G}^2$ under minimax optimization framework, while VGAN combining with BRC under hierarchical optimization framework is always able to obtain the optimal solution $\theta_{G}^1$ with the exact calculation of the corresponding constraint (i.e., BRC, $\theta_{D}(\theta_{G}^1)$).}
}\label{fig:pipeline}
\end{figure*}% 
\vspace{-0.4cm}	
\subsection{Our Motivations and Contributions} 
%We infer inappropriate modeling formulation and incorrect training techniques are two critical limiting factors. The main reason for arriving at this judgment is that the single-level minimax formulation and naive training strategy, for Vanilla GAN and its variants, cannot essentially depict the coupled hierarchical constraint relationships of the generator and discriminator networks. 
Despite the well-intended nature of Eq.~\eqref{eq:distance}, existing mathematical theory has proven that the convergence of minimax methods can be particularly fragile, even in simple, bilinear problems~\cite{daskalakis2021complexity}.  
Algorithmically, traditional GANs trained by taking simultaneous or alternating gradient updating rules (w.r.t. $\bm{\theta}_{D}$ and $\bm{\theta}_{G}$), lead to two separate optimization dynamics for generator and discriminator~\cite{mescheder2018training}. 
%They do not generally consider order aspects and thus cause various stability problems.
Both forms share the same drawbacks, stemming from the fact that they generally do not consider ordering aspects, thus causing various stability problems. This also means that the gradient updates of VGAN exhibit cyclic behavior, which may slow down convergence or cause phenomena such as mode collapse, mode hopping, etc~\cite{gemp2018global}. 

In Figure~\ref{fig:pipeline}, we provide an illustration in terms of the comparisons of modeling and solving strategy under different optimization frameworks (i.e., VGAN and GAN with BRC). It exhibits how to find the optimal solution from three sampled variables of generators (i.e., $\bm{\theta}_{G}^1$, $\bm{\theta}_{G}^2$, and $\bm{\theta}_{G}^3$) by performing VGAN (a) and VGAN with BRC (b), respectively.
The dotted box indicates the solution process by executing two methods separately. For VGAN, even for a locally optimal discriminator parameter $\theta_{D}^{*}$, only a suboptimal result $\theta_{G}^2$ can be found, i.e., 
$F(\bm{\theta}_{G}^3,\bm{\theta}_{D}^{*})\ge F(\bm{\theta}_{G}^1,\bm{\theta}_{D}^{*})\ge F(\bm{\theta}_{G}^2,\bm{\theta}_{D}^{*})$.  
This corresponds exactly to the fact that the alternate training strategy of VGAN will force the optimization of two networks into a non-correspondence state, thus always resulting in oscillating, unstable situations. 
On the contrary, our proposed framework calculates the corresponding BRC for three sampled variables of generator, that is, the optimal discriminator $\bm{\theta}_{D}(\bm{\theta}_{G})$ given $\bm{\theta}_{G}$, and then transports it to the generator based on the value-function $\varphi(\bm{\theta}_{G})$, and finally obtains the optimal solution $ \theta_{G}^1$, explained by $F(\bm{\theta}_{G}^3,\bm{\theta}_{D}(\bm{\theta}_{G}^3))\geq F(\bm{\theta}_{G}^2,\bm{\theta}_{D}(\bm{\theta}_{G}^2))\geq F(\bm{\theta}_{G}^1,\bm{\theta}_{D}(\bm{\theta}_{G}^1))$.  
Based on this analysis, it can be concluded that traditional single-level minimax based modeling and training techniques are two main limiting factors to relieve the unstable state of GANs, while our proposed BRC framework can essentially depict the hierarchical training dynamics and explicitly formulate the potential dependency of the generator on the discriminator. 
%Taken together, we infer inappropriate modeling formulation and incorrect training techniques are two critical limiting factors, that cannot essentially depict hierarchical training dynamics and coupling nested relationship of the generator and discriminator networks.  

In this paper, we provide a novel  reformulation and algorithm perspective to re-comprehend and interpret the intrinsic mechanism of GAN training procedure. 	
Specifically, we define a best-response constraint with respect to the discriminator maximization to depict the coupling nested relationship with respect to generator and discriminator, which is then incorporated into the divergence minimization procedure in terms of generator.  
%It exactly corresponds to a hierarchical optimization problem, which breaks away from the reliance on inherent single level minimax paradigm. 
Algorithmically, instead of using the traditional hierarchical updating rule, we construct a fast solution strategy from the perspective of implicit gradient with an outer-product-based Hessian approximation technique. Through this, unprecedented improvements in training stability and alleviation of mode collapse challenges can be achieved by a significant margin.  
	Our contributions can be summarized as follows:
\begin{itemize}
	\item 
	%By introducing a Best-Response Constraint (BRC) to precisely describe the one-to-one correspondence for the optimization process of divergence minimization, we spurn the inherent single level minimax optimization paradigm with a novel hierarchical perspective to revisit and reveal intrinsic principles of GAN and its variants.  
	Abandoning the inherent single-level minimax formulation, we introduce Best-Response Constraint (BRC), a novel hierarchical optimization  framework to revisit learning principles of GANs, which can explicitly formulate the potential dependency of the generator on the discriminator.
	%The optimization of discriminator as a follower is depicted through a nested best-response constraint, and then is dynamically back-propagated to the optimization process of the dominant generator. 
	%\item  We provide a general and flexible learning platform to revisit and reveal intrinsic principles of adversarial learning for GANs, through which most of current advanced GAN variants can achieve stable and quick convergence without introducing any architectural changes and loss choices. 
	\item To alleviate high computational complexity issues associated with traditional hierarchical optimization, we design an implicit gradient scheme with outer-product Hessian approximation as fast solution strategy, which is more computation-friendly and suitable for diverse high-dimensional large-scale real-world applications.  
	%The BRC paradigm can be served as a general and flexible learning platform through which most of the current GAN variants can be reformulated to achieve consistent performance gains without deliberately introducing any architectural adaptations and loss choices.
	%\item  From a unified perspective,  we provide a general learning platform through which to revisit and reveal the intrinsic principles of adversarial learning for GANs, through which most of existing GAN and its variants can be regarded as the simplified and exception versions that can be uniformly analyzed. 
	%\item Emphatically, as a general learning platform, our BRC methodology together with the proposed algorithmic scheme can be flexibly embedded into various existing GAN variants to uniformly improve them.  Abundant experiments demonstrate the superiority by a significant margin especially in terms of training stability and mode collapse. 
	\item As a general learning platform, our BRC methodology accompanied with the fast solution strategy can uniformly improve existing GAN variants  by a significant margin in a flexible manner.  Abundant  
	experimental results demonstrate the effectiveness and efficiency especially in training stability and mode collapse. 
	%\item  With the general and flexible BRC framework, we provide improvements on a variety of existing GANs and demonstrate consistent performance boosts without deliberate modifications like architecture adjustments and loss choices. 
	%Through this, unprecedented improvements can be achieved by a significant margin  in both synthetic and real-world datasets.
	% in terms of performance improvement, training stability and alleviation of mode collapse. 
	%\item To provide a general learning platform and a flexible insight, we experimentally demonstrated BRC together with the proposed solution strategy can be embedded in most of current advanced GAN variants without introducing any architectural changes and loss choices, maintaining more stable training performance, fast convergence capability and better generalization capabilities in both synthetic and real-world areas. 
\end{itemize}
\vspace{-0.2cm} 
\section{The Proposed Framework} 
In this section, we introduce Best-Response Constraint (BRC) to accurately portray the coupling relationship, and then propose a fast solution strategy with implicit gradient approximation. We also explore how to uniformly improve existing GAN variants under BRC platform.
%we propose the best-response constraint to accurately and explicitly describe the coupling relationship between the generator and the discriminator, to break the normal unstable state under the original training mechanism. 
%Essentially it can be understood as incorporating adversarial training as a constraint into a hierarchical leader-follower learning framework, replacing the optimization of minimax. 
\vspace{-0.2cm} 
\subsection{Adversarial Learning with Best-Response Constraints}
\vspace{-0.2cm}
As the single-level type modeling formulation of GAN have proven to be still notoriously unstable, we seek to construct a novel hierarchical learning perspective to portray the $``\mathtt{distance}"$  in Eq.~\eqref{eq:distance}.
Starting from maximization optimization of discriminator, we define a BRC, i.e., $\bm{\theta}_{D}(\bm{\theta}_{G})\in\mathcal{S}(\bm{\theta}_{G})$ to govern the dynamic update process of the divergence minimization of generator.
Given a generator with parameter $\bm{\theta}_{G}$, the BRC can be formulated as the optimization of discriminator network followed by a process of maximizing optimization, 
\begin{equation}
	\mathcal{S}(\bm{\theta}_{G}):=\arg\max\limits_{\bm{\theta}_{D}}f(\bm{\theta}_{D},\bm{\theta}_{G}).\label{eq:lower-level}
\end{equation}
$\mathcal{S}(\bm{\theta}_{G})$ denotes the solution set  with respect to any given $\bm{\theta}_{G}$.  
It can be intuitively interpreted as the training of the discriminator parameters until optimal when current $\bm{\theta}_{G}$ is fixed. 	
The concept of an optimal discriminator is particularly important because its optimality would ensure that the discriminator produces meaningful feedback to the generator. 	
Then by enforcing the optimal discriminator 
$\bm{\theta}_{D}(\bm{\theta}_{G})$ as a best-response to the divergence objective, we define the value-function $\varphi(\bm{\theta}_{G})$ to more accurately depict the divergence minimization problem, i.e.,
\begin{equation}
	\min\limits_{\bm{\theta}_{G}}\varphi(\bm{\theta}_{G}):=F(\bm{\theta}_{G},\bm{\theta}_{D}(\bm{\theta}_{G})).\label{eq:mini-varphi}
\end{equation}
Intuitively, the above process contains two levels of optimization tasks where the optimization of discriminator as a constraint is nested within the generative learning task.  
From this perspective,  generative adversarial learning corresponds to a hierarchical leader-follower optimization problem 	
\begin{equation}
	\min\limits_{\bm{\theta}_{G}}F(\bm{\theta}_{G},\bm{\theta}_{D}(\bm{\theta}_{G})),\ s.t. \ \bm{\theta}_{D}(\bm{\theta}_{G})\in \mathcal{S}(\bm{\theta}_{G}),\label{eq:bilevelgan}
\end{equation} 
where the leader objective $F$ tries to minimize the expected divergence between characteristic functions of real and generated data distributions, the maximizition of the follower objective $f$ targets at obtaining the optimal discriminator transmitted to the generator divergence. As illustrated in Figure~\ref{fig:pipeline}, for any generator $G$, there is a corresponding (parametric) discriminator $D$ in the follower problem to be learned that provides the rational (optimal) response to the leader generator. 
It is not symmetric in terms of two levels and the discriminator $D$ is served as a constraint on the generator $G$ to optimize $\bm{\theta}_{G}$, which is in accordance with the intrinsic mechanism of GANs as a generative framework.
%~\cite{mescheder2018training}.  

Indeed, the gradient of best-response constraint (i.e., $\frac{\partial \bm{\theta}_{D}(\bm{\theta}_{G})}{\partial \bm{\theta}_{G}}$) captures how the optimal discriminator $\bm{\theta}_{D}(\bm{\theta}_{G})$ responds to the change in generator. 
Thus, the optimization (w.r.t. $\bm{\theta}_{G}$)  for the divergence minimization problem can be further derived as 
\begin{equation}
	\frac{\partial \varphi(\bm{\theta}_{G})}{\partial\bm{\theta}_{G}}=\overbrace{\frac{\partial F(\bm{\theta}_{G},\bm{\theta}_{D})}{\partial \bm{\theta}_{G}}}^{\mbox{Direct-Response Gradient}} + \overbrace{\frac{\partial F(\bm{\theta}_{G},\bm{\theta}_{D})}{\partial \bm{\theta}_{D}}\underbrace{\frac{\partial \bm{\theta}_{D}(\bm{\theta}_{G})}{\partial \bm{\theta}_{G}}}_{\mbox{Best-Response (BR)  Gradient}}}^{\mbox{Hierarchical Coupled-Response Gradient}}.\label{eq:bilevel-gradient}
\end{equation}
Here, the BR gradient term denotes the gradient of best-response constraint, and the direct-response gradient term reveals straightforward dependence on generator. 
The intuition is that capturing how a discriminator would react given a change in discriminator and feeding it into a generator would allow the generator to react, and prevent mode collapse.  More specifically, the response gradient term has the ability to embed rich adversarial information learned from the lower-level discriminator $D$ to promote the propagation of the generator $G$. 
Indeed, the principle mechanism inherent of GANs can be further understood, i.e., the optimal feedback to the generator $G$ is acquired by multiple updates to the discriminator $D$. 
Thus the BR Gradient term of the follower actually plays the key role for the adversarial learning process.

However, as for most of existing GANs, they just establish gradient flows simply based on the direct-response gradient term of Eq.~\eqref{eq:bilevel-gradient} to update  $\bm{\theta}_{D}$ and $\bm{\theta}_{G}$, i.e., $\bm{\theta}_{D}^{t+1}=\bm{\theta}_{D}^t+ s_t \frac{\partial  F \left(\bm{\theta}_{G}^t,\bm{\theta}_{D}^t \right)}{\partial \bm{\theta}_{D}^t},$ and $\bm{\theta}_{G}^{t+1}=\bm{\theta}_{G}^t- l_t \frac{\partial  F \left(\bm{\theta}_{G}^t,\bm{\theta}_{D}^t \right)}{\partial \bm{\theta}_{G}^t}$  ($s_t $ and $l_t $ represent the learning rates for $ G $ and $ D $ updates), which either ignores BR gradient or inaccurately approximates best-response without unrolled gradient.  
Thus all fail to exactly evaluate the complicated dependency of the generator induced by $\bm{\theta}_{D}(\bm{\theta}_{G})$. It can be directly reflected by why the optimal solution cannot be found by VGAN even for a locally optimal discriminator parameter $\theta_{D}^{*}$, as illustrated in (a) of Figure~\ref{fig:pipeline}. 
Explained from this perspective, our proposed BRC framework has the ability to better capture the potential dependency of the generator on the discriminator, thus maintaining better generalization capabilities, refer to (b) in Figure~\ref{fig:pipeline}.  
\subsection{Fast Solution Strategy with Implicit Gradient Approximation} 
In this section, we propose a fast and efficient training strategy, Implicit Gradient Approximation (IGA), to alleviate the complexity and instability of adversarial learning process. 
The key step in the preceding discussion is providing the accurately calculated  BR gradient (i.e., $\frac{\partial \bm{\theta}_{D}(\bm{\theta}_{G})}{\partial \bm{\theta}_{G}}$ of Eq.~\eqref{eq:bilevel-gradient}.
However, it is computationally challenging for most existing strategies to evaluate the exact BR gradient. This difficulty is further aggregated when $\bm{\theta}_{D}$ and $\bm{\theta}_{G}$ are high-dimensional. 
Considering that implicit methods can directly and precisely estimate the optimal gradient, we derive the following equation in the case of optimal discriminator, i.e., $ \partial f/ \partial\bm{\theta}_{D}=0$, inspired by implicit function theorem. 
Thus, the BR gradient that is inextricably linked with discriminator maximization is then replaced with an implicit equation, i.e., 
\begin{equation}
	\frac{\partial \bm{\theta}_{D}(\bm{\theta}_{G})}{\partial \bm{\theta}_{G}} 
	= -\left( \frac{\partial^2 f(\bm{\theta}_{G},\bm{\theta}_{D}(\bm{\theta}_{G}))}{\partial \bm{\theta}_{D} \partial \bm{\theta}_{D}} \right) ^{-1}
	\frac{\partial^2 f(\bm{\theta}_{G},\bm{\theta}_{D}(\bm{\theta}_{G}))}{\partial \bm{\theta}_{D} \partial \bm{\theta}_{G} }. \label{eq:ift}
\end{equation} 

On account of the extreme difficulty of counting Hessian and its inverse, we seek to establish an efficient solving strategy from the perspective of simplification of second derivative to first derivative to compute the best-response Jacobian, accompanied by two key computational steps, implicit gradient estimation and outer-product approximation.

\textbf{Implicit Gradient Estimation.} For convenience,  we first denote the hierarchical coupled-response gradient term in Eq.~\eqref{eq:bilevel-gradient} as $\mathbf{G}_{R}$, i.e.,  $\mathbf{G}_{R}=\frac{\partial F(\bm{\theta}_{G},\bm{\theta}_{D})}{\partial \bm{\theta}_{D}}\frac{\partial \bm{\theta}_{D}(\bm{\theta}_{G})}{\partial \bm{\theta}_{G}}.$   To avoid directly calculating the products of various Hessians and their inversions, we further introduce a linear solver system $\mathbf{B}$ based on Eq.~\eqref{eq:ift}, for evading the complexity of calculating $\mathbf{G}_{R}$.
Hence, the indirect response gradient can be reformulated as
\begin{equation}
	\mathbf{G}_{R}=\left(\frac{\partial^2f}{\partial\bm{\theta}_{D}\partial\bm{\theta}_{G}}\right)^{\top}\mathbf{B}, \ \mbox{where} \ \frac{\partial^2f}{\partial\bm{\theta}_{D}\partial\bm{\theta}_{D}}\mathbf{B}=-\frac{\partial F}{\partial \bm{\theta}_{D}},\label{eq:adjoint-equation}
\end{equation}
where $(\cdot)^{\top}$ denotes the transposition operation. 	
The indirect response gradient $\mathbf{G}_{R}$ only depends on the first-order condition, and decouples the computational burden from the solution trajectory of the lower-level discriminator, which can greatly reduce the pressure to propagate the backward gradient in the discriminator maximization dynamics. 
However, it can be recognized that the calculation of second-order derivatives in $\mathbf{G}_{R}$ is still intractable. 
The urgent requirements from  approximating the repeated computation of two Hessian matrix $\frac{\partial^2f}{\partial\bm{\theta}_{D}\partial\bm{\theta}_{D}}$ and  $\frac{\partial^2f}{\partial\bm{\theta}_{D}\partial\bm{\theta}_{G}}$  spawned the following concept of outer-product approximation.

\textbf{Outer-Product Approximation.} To further suppress the complexity of discriminator optimization, we consider replacing the original Hessian operation based on Gauss-Newton method and introduce two approximations using corresponding outer products, as follows:
\begin{equation}
	\frac{\partial^2f}{\partial\bm{\theta}_{D}\partial\bm{\theta}_{D}} \approx \frac{\partial f}{\partial\bm{\theta}_{D}}\frac{\partial f^{\top}}{\partial\bm{\theta}_{D}},\
	\frac{\partial^2f}{\partial\bm{\theta}_{D}\partial\bm{\theta}_{G}} \approx \frac{\partial f}{\partial\bm{\theta}_{D}}\frac{\partial f^{\top}}{\partial\bm{\theta}_{G}}.\label{eq:outer-product}
\end{equation}
This way of gradient separation converts the second-order derivative of high complexity into a simple product operation of the first-order derivative, which can greatly reduce the complexity of the algorithm, especially in the memory consumption. 
By combining Eqs.~\eqref{eq:adjoint-equation}-\eqref{eq:outer-product}, we establish the nonlinear least squares problem by approximating the Gauss-Newton formula. Plugging into the Eq.~\eqref{eq:adjoint-equation}, thus we can obtain  $\mathbf{B}\approx -\left(\left(\frac{\partial f}{\partial\bm{\theta}_{D}}\frac{\partial f^{\top}}{\partial\bm{\theta}_{D}}\right)^{\top}\frac{\partial f}{\partial\bm{\theta}_{D}}\frac{\partial f^{\top}}{\partial\bm{\theta}_{D}}\right)^{-1}\left(\frac{\partial f}{\partial\bm{\theta}_{D}}\frac{\partial f^{\top}}{\partial\bm{\theta}_{D}}\right)^{\top}\frac{\partial F}{\partial \bm{\theta}_{D}}.$ \vspace{-0.1cm}
To over simplify,  we can have $\mathbf{B}=\left(\frac{\partial f}{\partial\bm{\theta}_{D}}\frac{\partial f^{\top}}{\partial\bm{\theta}_{D}}\frac{\partial f}{\partial\bm{\theta}_{D}}\frac{\partial f^{\top}}{\partial\bm{\theta}_{D}}\right)^{-1}\left(\frac{\partial f^{\top}}{\partial\bm{\theta}_{D}}\frac{\partial f}{\partial\bm{\theta}_{D}}\right)\frac{\partial F}{\partial \bm{\theta}_{D}}.$
%\[
%\begin{aligned}		 
%	\mathbf{B}& \approx -\left(\left(\frac{\partial f}{\partial\bm{\theta}_{D}}\frac{\partial f^{\top}}{\partial\bm{\theta}_{D}}\right)^{\top}\frac{\partial f}{\partial\bm{\theta}_{D}}\frac{\partial f^{\top}}{\partial\bm{\theta}_{D}}\right)^{-1}\left(\frac{\partial f}{\partial\bm{\theta}_{D}}\frac{\partial f^{\top}}{\partial\bm{\theta}_{D}}\right)^{\top}\frac{\partial F}{\partial \bm{\theta}_{D}} \\
%	& = -\left(\frac{\partial f}{\partial\bm{\theta}_{D}}\frac{\partial f^{\top}}{\partial\bm{\theta}_{D}}\frac{\partial f}{\partial\bm{\theta}_{D}}\frac{\partial f^{\top}}{\partial\bm{\theta}_{D}}\right)^{-1}\left(\frac{\partial f^{\top}}{\partial\bm{\theta}_{D}}\frac{\partial f}{\partial\bm{\theta}_{D}}\right)\frac{\partial F}{\partial \bm{\theta}_{D}}. \\
%\end{aligned}
%\]
Finally, the representation of the response gradient $\mathbf{G}_R$ is approximately obtained, i.e.,
\begin{equation}
	\mathbf{G}_R\approx-\frac{\partial f}{\partial \bm{\theta}_{G}}\left({\frac{\partial f^{\top}}{\partial \bm{\theta}_{D}}\frac{\partial F}{\partial \bm{\theta}_{D}}}\bigg/{\frac{\partial f^{\top}}{\partial\bm{\theta}_{D}}\frac{\partial f}{\partial\bm{\theta}_{D}}}\right).\label{eq:approximate-G}
\end{equation}	
\begin{wrapfigure}{r}{8cm}
	\vspace{-0.7cm} %
	\begin{minipage}{1.0\linewidth}
		\begin{algorithm}[H]
			\caption{Implicit Gradient Approximation (IGA).}\label{alg:fast}
			\begin{algorithmic}[1]
				\REQUIRE Initialization $\bm{\theta}_G^0, \bm{\theta}_D^0$ and necessary parameters. 
				\ENSURE  The optimal generator $G$.
				\WHILE{not converge}
				\STATE Obtain approximation $\hat{\bm{\theta}}_D$ to $\bm{\theta}_D$ by solving Eq.~\eqref{eq:lower-level}.  
				\vspace{-0.25cm}	
				\STATE Calculate $\mathbf{G}_R$ with $\hat{\bm{\theta}}_D$ and current $\bm{\theta}_G$ by Eq.~\eqref{eq:approximate-G}.
				\STATE Calculate $\frac{\partial \varphi(\bm{\theta}_{G})}{\partial\bm{\theta}_{G}}$ by Eq.~\eqref{eq:bilevel-gradient} .
				\STATE Update $\bm{\theta}_G \leftarrow \bm{\theta}_G - \alpha\frac{\partial \varphi(\bm{\theta}_{G})}{\partial\bm{\theta}_{G}}$ ($\alpha$: learning rate).
				\ENDWHILE
				\RETURN The optimal parameters $\bm{\theta}_G$.
			\end{algorithmic}
		\end{algorithm}
	\end{minipage}
	\vspace{-0.3cm}
\end{wrapfigure} 
Based on the above derivations, we can summarize the complete solution strategy in Alg.~\ref{alg:fast}.  In the training process, with the current generator parameters $\bm{\theta}_{G}$, we first optimize the discriminator according to the discriminator maximization objective $f$ in several steps to approximate the optimal discriminator, i.e., $\hat{\bm{\theta}}_D(\bm{\theta}_{G}) \approx \bm{\theta}_D(\bm{\theta}_{G})$. The optimal discriminator is then back-propagated to the generator, $\mathbf{G}_R$ is calculated based on our proposed implicit gradient strategy (i.e., Eq.~\eqref{eq:approximate-G}), which in turn updates the parameters of the generator until convergence.

In terms of computational complexity,  Alg.~\ref{alg:fast} does not involve the unfolded dynamical system by recurrent iteration or Hessian inverse, without any computation of Hessian- or Jacobian-vector products. All the complexity of our strategy comes from computing the first order gradient.	
The calculation of the first derivative of the function and the Hessian vector product have similar time and space complexity, our approximate method simplifies the process of solving the response gradient $\mathbf{G}_R$ to only calculate the first-order derivative of a few fixed times. We also provide a fair assessment of the various traditional hierarchical methods in Table~\ref{tab:time2}, verifying the superiority of our solution strategy in terms of convergence speed and memory consumption.
\vspace{-0.2cm}
\subsection{BRC vs. Existing GAN variants}
\vspace{-0.2cm} 
\textbf{Issues of Existing GAN variants.} In recent years, a plethora of GAN variants have been established to improve performance, from utilizing better loss functions (e.g., WGAN~\cite{arjovsky2017wasserstein}, LSGAN~\cite{mao2017least}, SNGAN~\cite{miyato2018spectral}) to designing suitable training strategies (e.g., UnrolledGAN~\cite{metz2016unrolled}, ProxGAN~\cite{farnia2020gans}). Even with different formulations and  complex strategies, these methods are always notoriously fragile and unstable in most cases. For these GAN methods, the core problem is that the generator and discriminator are modeled and optimized through the same objective $F$ based on single-level minimax type formulation, and then perform the alternate propagation strategy under two parallel patterns. With this setting, the optimization of generator always depends on parameters of discriminator in the previous step rather than the current step, as illustrated in (a) of Figure~\ref{fig:pipeline}. This directly results in deviating from the essential goal of divergence minimization process in Eq.~\eqref{eq:distance}, and triggers unstable situations like mode collapse and gradient disappearance.  

\textbf{Uniformly Improving GAN Variants by BRC.}  Despite different motivations and formulations, all existing GANs can be uniformly improved under our flexible BRC platform. With BRC methodology, we can first construct the discriminator dynamics corresponding to the maximization optimization of Eq.~\eqref{eq:lower-level}. As illustrated in Figure~\ref{fig:pipeline} from (a) to (b), the optimization of discriminator can be depicted by an exact estimated BRC. Then the BRC can be dynamically back-propagated to the optimization process of Eq.~\eqref{eq:bilevelgan} for the generator dynamics. Thus, it can accurately support the solving procedure of divergence minimization dynamics between generator and discriminator for all GANs. With the proposed fast solution strategy, we compute the hierarchical response gradient for generator with the outer-product approximation via Eq.~\eqref{eq:adjoint-equation} and Eq.~\eqref{eq:outer-product}. Through this, the potential dependency of the generator on the current discriminator can be accurately formulated and well-portrayed.  We demonstrate the effectiveness and efficiency of BRC  in terms of more stable training performance and better generalization capabilities through abundant experimental results.  

\vspace{-0.1cm}
\section{Experimental Results} \label{exp}
\vspace{-0.1cm}	
In this section, we conduct comprehensive experiments on the  synthetic and real-world applications to evaluate the effectiveness and flexibility of our proposed BRC framework and solution strategy.
\vspace{-0.2cm}
\subsection{Evaluations on Synthesized Data}
\vspace{-0.15cm}
\begin{wrapfigure}{r}{8.4cm}
	\vspace{-0.6cm}
	\begin{minipage}[t]{1.0\linewidth}
		\centering 
		\makeatletter\def\@captype{table}\makeatother \caption{ Comparison results on three representative GAN methods on 2D Ring MOG experiment. Best and second best results are \textbf{highlighted} and \underline{underlined}.} \vspace{0.1cm}
		\setlength{\tabcolsep}{1.8mm}{
			\begin{tabular}{|c|c|c|c|}
				\hline
				 \multirow{2}[0]{*}{\footnotesize Method} & \multicolumn{3}{c|}{2D Ring (Max Mode=5)} \\
				 \cline{2-4}		
				 & \footnotesize  FID$\downarrow$  & \footnotesize  JS$\downarrow$  & \footnotesize Mode$\uparrow$ \\
				\hline 
				\hline
				\footnotesize VGAN  &  \footnotesize 295.5$\pm$16.20  &  \footnotesize 0.798 $\pm$0.08 &  \footnotesize 2.54$\pm$0.51\\
				\hline 
				\footnotesize WGAN  & \footnotesize 28.93$\pm$4.10  & \footnotesize 0.728$\pm$0.07 & \underline{\footnotesize 3.80$\pm$1.50}\\
				\hline 
				\footnotesize UnrolledGAN  & \footnotesize 246.9$\pm$7.60  & \footnotesize 0.796$\pm$0.02 & \footnotesize 1.90$\pm$1.26\\
				\hline 
				\footnotesize VGAN(BRC)  & \underline{ \footnotesize 15.43$\pm$0.77 } & \underline{\footnotesize 0.60$\pm$0.27  } & \textbf{\footnotesize 5.00$\pm$0 } \\
				\hline 
				\footnotesize VGAN(BRC-R$_f$)  & \textbf{ \footnotesize 0.87$\pm$0.07 } &\textbf{\footnotesize  0.39$\pm$0.31 } & \textbf{ \footnotesize 5.00$\pm$0 }\\
				\hline			
			\end{tabular}%
		}	 
		\label{tab:ring5}%
	\end{minipage}
	\vspace{-0.4cm} 
\end{wrapfigure}
\textbf{Implementation Details.} First, on the synthetic dataset, we discuss the feasibility of the proposed algorithm in terms of operational efficiency and convergence performance compared with the current mainstream GAN networks.
To fully demonstrate the effectiveness of our method, we perform visual analysis and comprehensive performance evaluation on 2D and 3D Mixed of Gaussian (MOG) distribution data. We compare performance with several state-of-the-art GAN architectures that mitigate mode collapse and maintain stable training, including VGAN~\cite{goodfellow2014generative}, WGAN~\cite{arjovsky2017wasserstein}, ProxGAN~\cite{farnia2020gans}, LCGAN~\cite{engel2017latent}, etc. 

In the 2D case, we create 2D Ring MOG distribution with 5 or 8 2D Gaussian functions  (denoted by ``Max Mode=5/8''). In the 3D scene, we conduct this experiment with 27 3D Gaussian functions in a cube (denoted by ``Max Mode=27''). The variance of each Gaussian distribution is set to 0.02. During the training phase, we create 512 generated data and real data samples from each mixture of Gaussian models for each training batch, and sample 512 generated images for testing. 
The Adam optimizer is uniformly set to optimize two networks, and the learning rates of the discriminator $D$ and generator $G$ are $10^{-4}$ and $10^{-3}$, respectively. For the generator and discriminator, we use a 3-layer linear network with a width of 256 and a leaky relu with a threshold of 0.2 as the activation function.
We employ three well-known metrics, i.e.,  Frechet Inception Distance (FID)~\cite{heusel2017gans}, Jensen-Shannon divergence (JS)~\cite{goodfellow2014generative}, number of Modes (Mode), to provide a comprehensive comparison. 
It should be noted that in the synthetic experiments, we directly compute the data itself as a feature to measure FID, which requires the use of a pretrained network for feature extraction.

\textbf{Performance Evaluation.} On 2D Ring MOG experiment,  we first evaluate the effectiveness and efficiency of our method for the improvement of VGAN, and demonstrate the flexibility in choosing lower-level loss function under our framework. As shown in Table~\ref{tab:ring5},  compared to WGAN and UnrolledGAN, original VGAN combining with BRC is distinguished by immensely increased significant performance improvement with better score (FID, JS and Mode) and achieves comparable training efficiency in speed. 
On this basis, we introduce an extra regular term  for the discriminator based on VGAN, denoted by BRC-R$_f$, which can further stabilize training and improve performance. 
In this way, the discriminator loss function can be expressed as  
$	 
f(\bm{\theta}_G,\bm{\theta}_D) =f_1(\bm{\theta}_G,\bm{\theta}_D)+f_2(\bm{\theta}_G,\bm{\theta}_D),
$
where  $f_1(\bm{\theta}_G,\bm{\theta}_D)=E_{\mathbf{x}\sim P_{data}}\left[\log(D_2(D_1(\mathbf{x})),\mathtt{mean}\left\{D_1(\mathbf{x}),\mathtt{var}(D_1(\mathbf{x}))\right\}\right],$ 
$
f_2(\bm{\theta}_G,\bm{\theta}_D) =E_{\mathbf{z}\sim P_{\mathbf{z}}}\left[\log(1-D_2(D_1(G(\mathbf{z})),h
(D_1)))\right],  
$
and $h(D_1)=\mathtt{mean}\left\{D_1(G(\mathbf{z})), \mathtt{var}(D_1(G(\mathbf{z})))\right\}$. Here,
$D=D_2\circ D_1$, $D_1$ and $D_2$ are the feature extraction layer and  the linear classification layer of $D$, respectively. That is, we add the mean and variance of data features as input to the discriminator classification layer. Thanks to the flexibility of our BRC framework to revise the discriminator objective function, VGAN can achieve optimal performance.
\begin{wrapfigure}{r}{8.5cm}
	\vspace{-0.3cm} 
	\setlength{\tabcolsep}{1.5mm}{
		\begin{tabular}{r@{\extracolsep{0.1em}}r@{\extracolsep{0.1em}}r}
			\includegraphics[height=3.5cm]{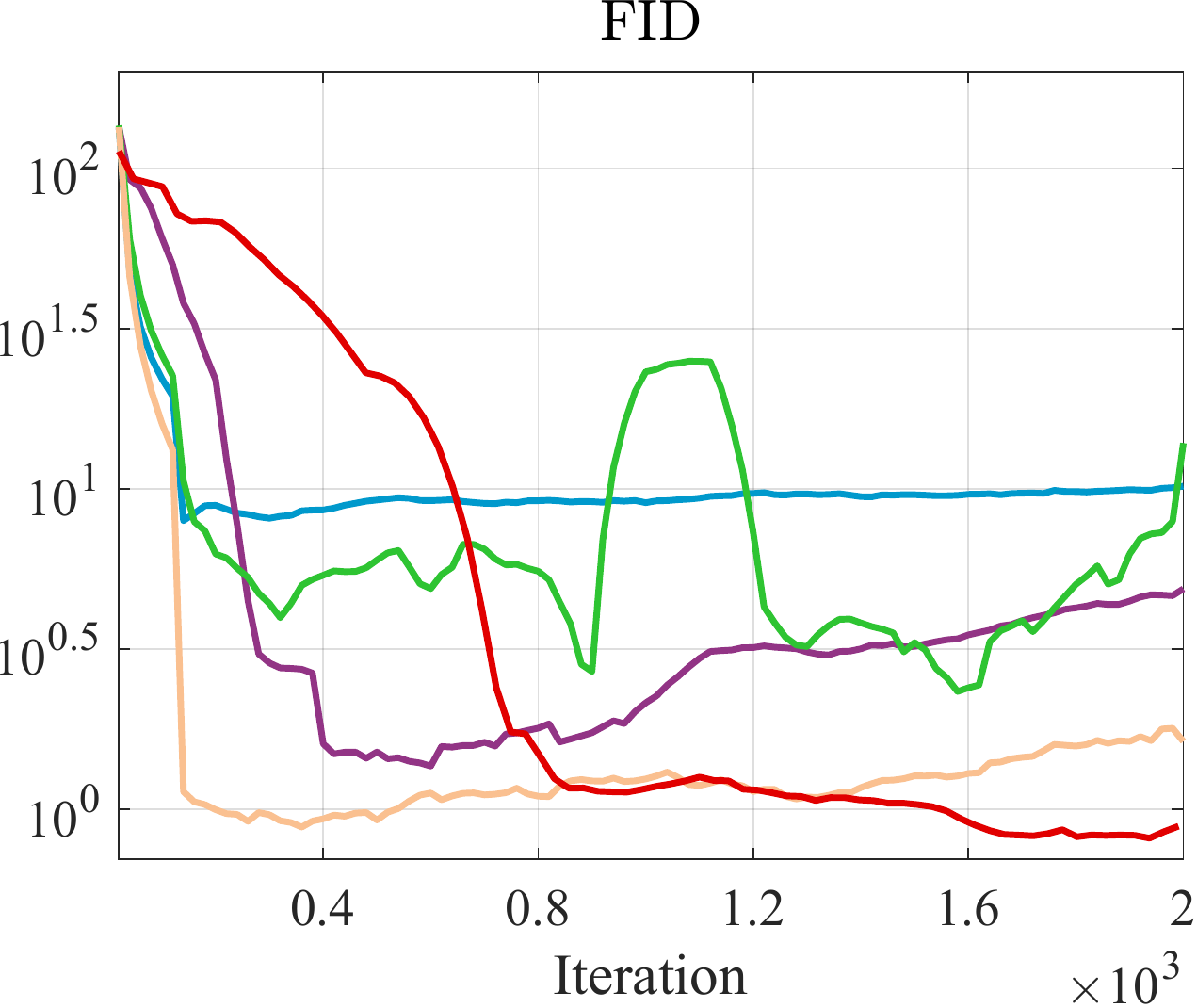}&&\includegraphics[height=3.5cm]{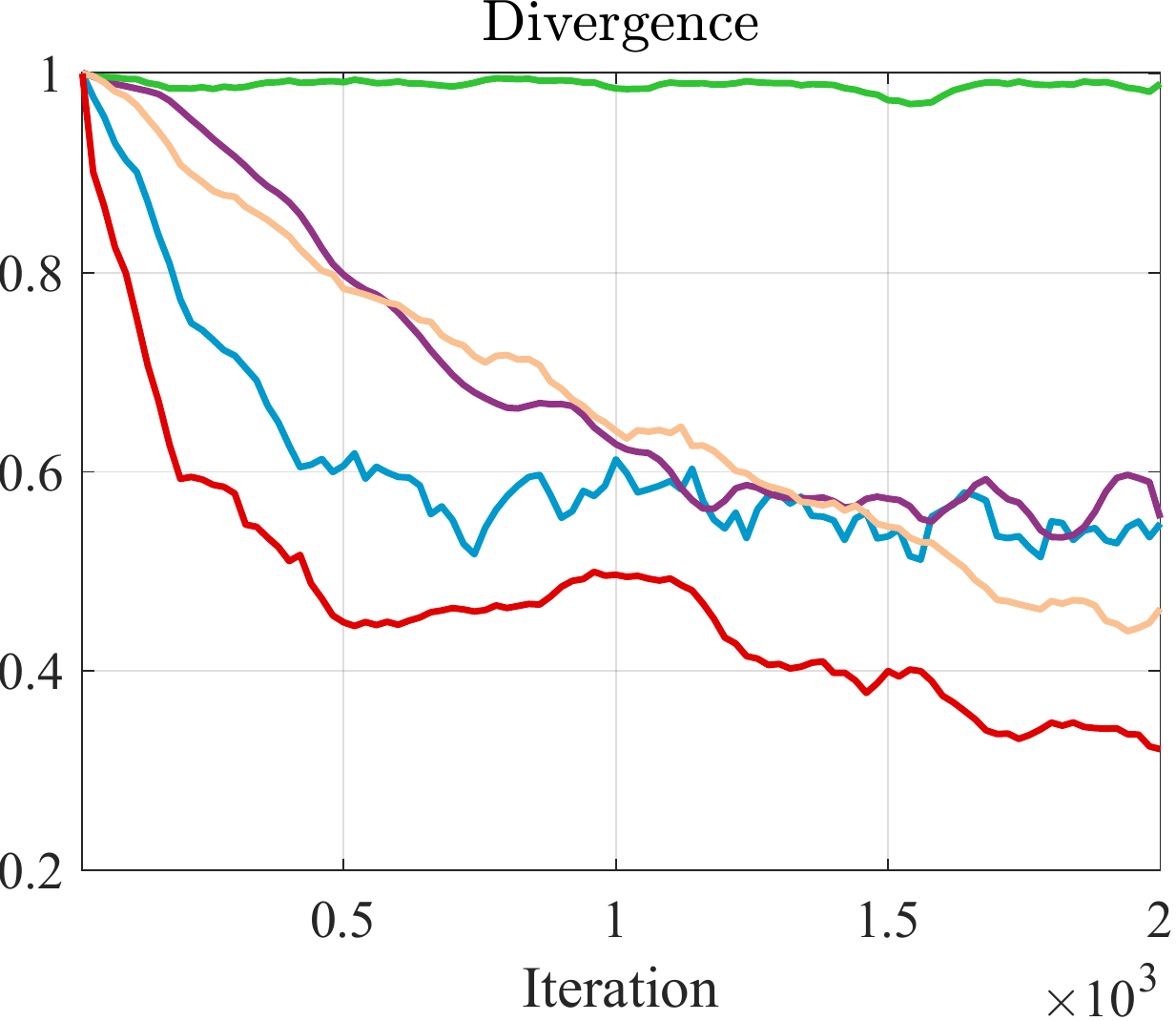}\\
			\multicolumn{3}{r}{\includegraphics[width=0.95\linewidth,trim=0 0 0 0,clip]{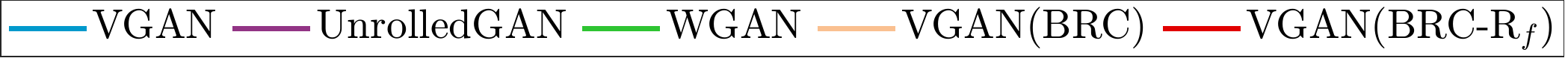}}\\
			%\multicolumn{3}{c}{\includegraphics[width=0.95\linewidth,trim=0 0 0 300,clip]{fig/js.pdf}}\\
		\end{tabular}
	} \vspace{-0.2cm}	
	\caption{Comparison results of FID and JS at each training iteration in 2D Ring MOG experiment. To improve the convergence of the algorithm, an additional rule term (denoted as BRC-R$_f$) is introduced in the discriminator loss of VGAN.}
	\label{fig:syn_convergence}
	\vspace{-0.5cm}
\end{wrapfigure}
%\begin{wrapfigure}{r}{8.5cm}
%	\vspace{-0.3cm} 
%	\setlength{\tabcolsep}{1.5mm}{
%		\begin{tabular}{c@{\extracolsep{0.1em}}}
%			\includegraphics[width=0.98\linewidth]{fig/JS3.pdf}\\
%		\end{tabular}
%	}	
%	\caption{Comparison results of FID and JS at each training iteration in 2D Ring MOG experiment. To improve the convergence of the algorithm, an additional rule term (denoted as BRC-R$_f$) is introduced in the discriminator loss of VGAN.}
%	\label{fig:syn_convergence}
%	\vspace{-0.5cm}
%\end{wrapfigure}
Figure~\ref{fig:syn_convergence} illustrates the FID and JS curves in terms of various methods (i.e., VGAN, WGAN, UnrolledGAN, VGAN(BRC) and VGAN(BRC-R$_f$)). It can be seen that VGAN(BRC-R$_f$) and VGAN(BRC) can finally achieve the best and second best performance metrics.

\begin{figure*}[!t]
	\vspace{-0.4cm}
	\begin{center}
		\begin{tabular}{c@{\extracolsep{0.4em}}c@{\extracolsep{0.4em}}c@{\extracolsep{0.4em}}c@{\extracolsep{0.4em}}c@{\extracolsep{0.4em}}c@{\extracolsep{0.4em}}c@{\extracolsep{0.4em}}}
			\hline \toprule[0.65pt] \specialrule{0em}{1pt}{1pt} 
			\multirow{5}{*}{\footnotesize \textbf{Target}} & \footnotesize \textbf{VGAN} & \footnotesize \textbf{WGAN} & \footnotesize \textbf{ProxGAN} & \footnotesize \textbf{LCGAN} & & \multirow{7}{*}{\textbf{w/o}}\\
			\vspace{-0.4cm}
			\multirow{10}{*}{\includegraphics[height=2.2 cm,trim=0 0 0 0,clip]{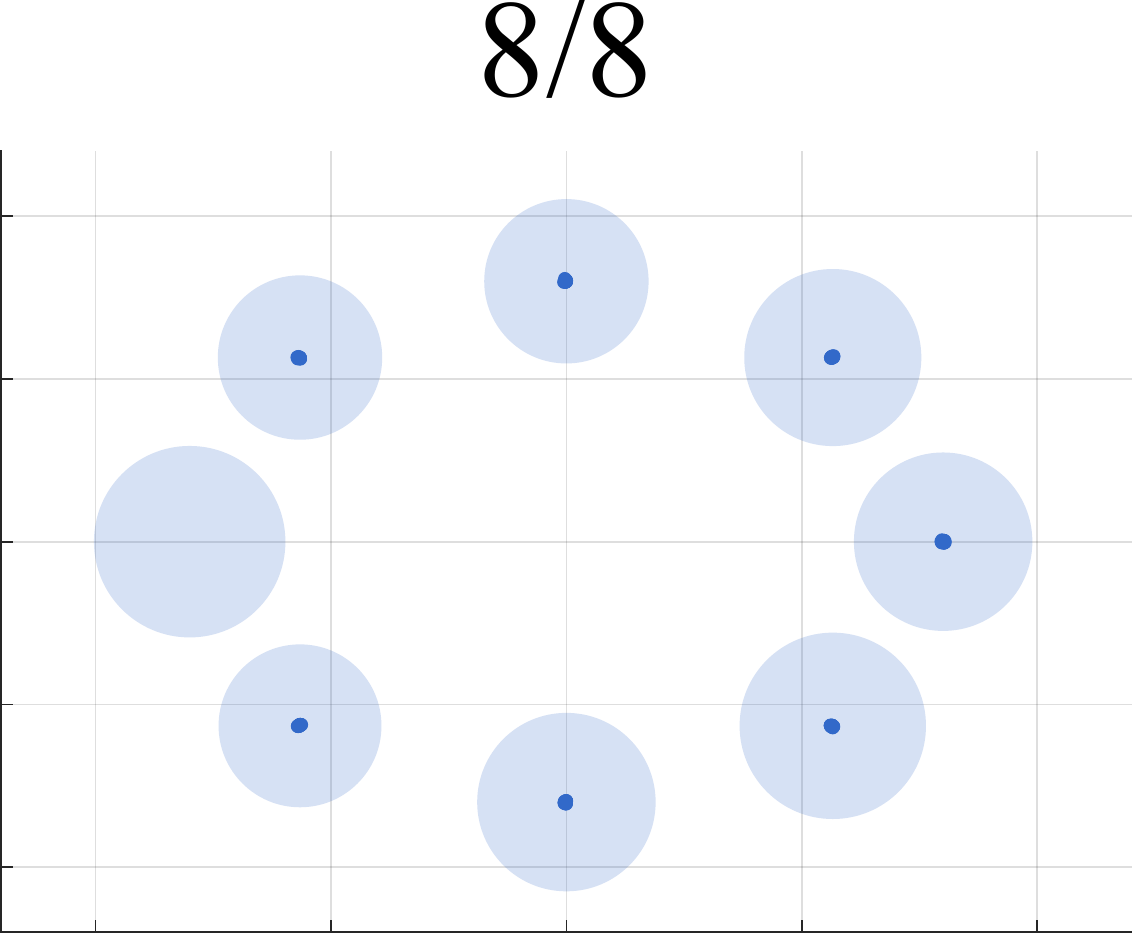}}& &&& & & \\
		 &\includegraphics[height=1.9cm]{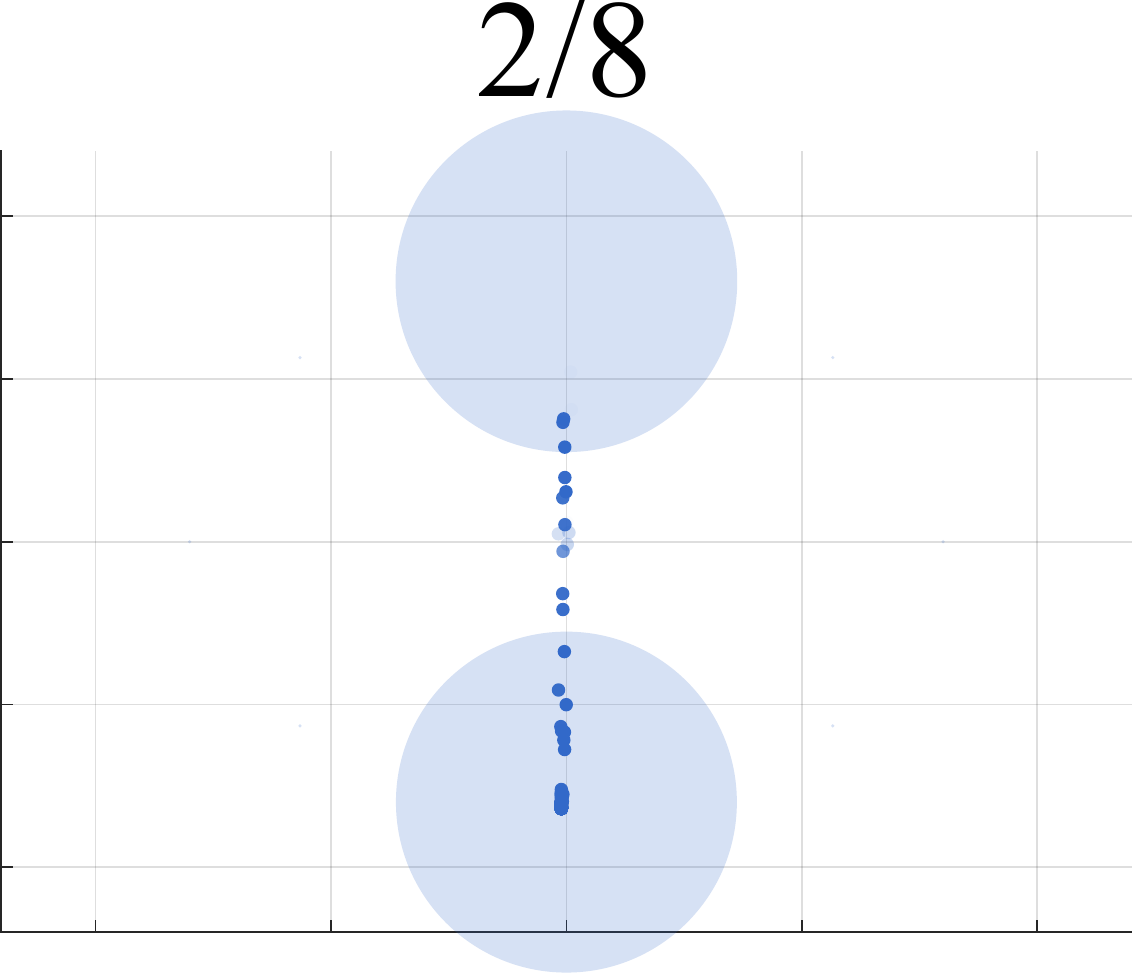}& \includegraphics[height=1.9cm]{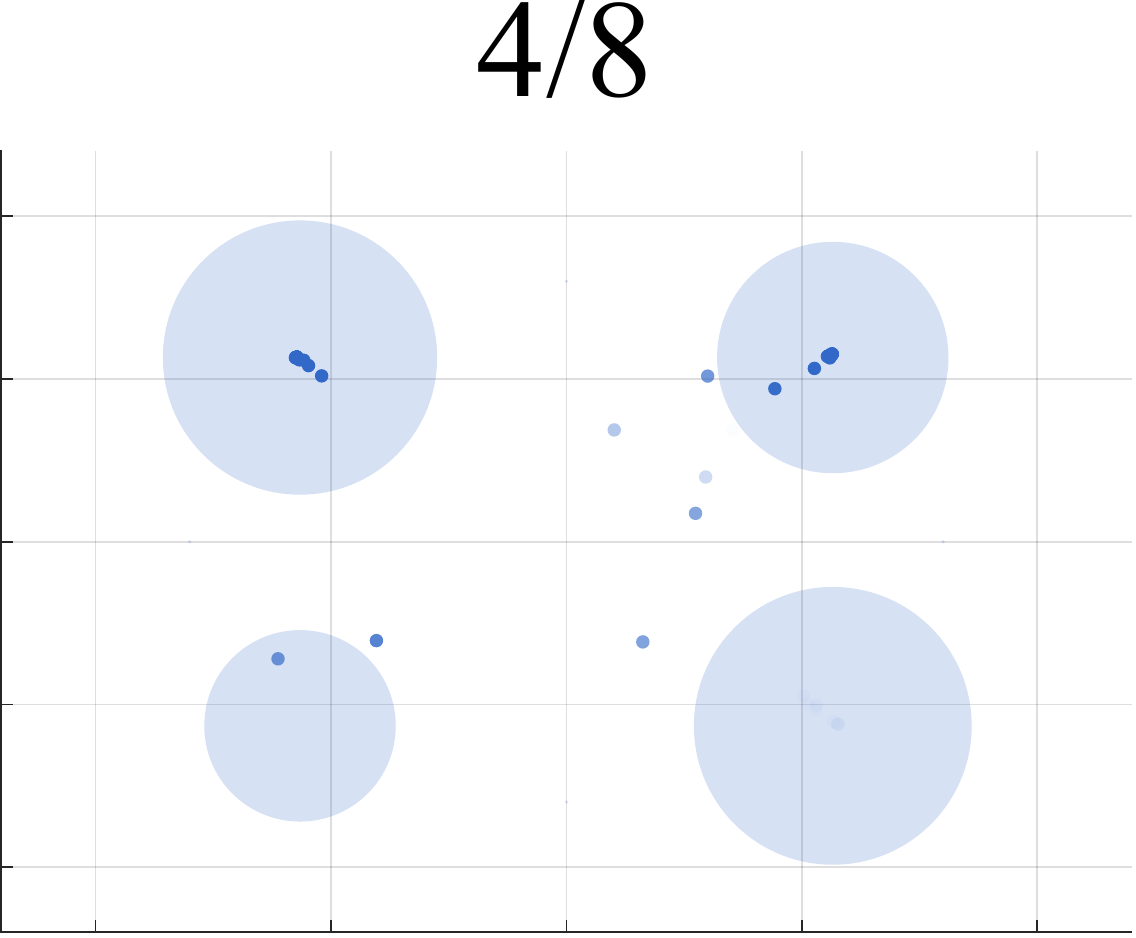} & \includegraphics[height=1.9cm]{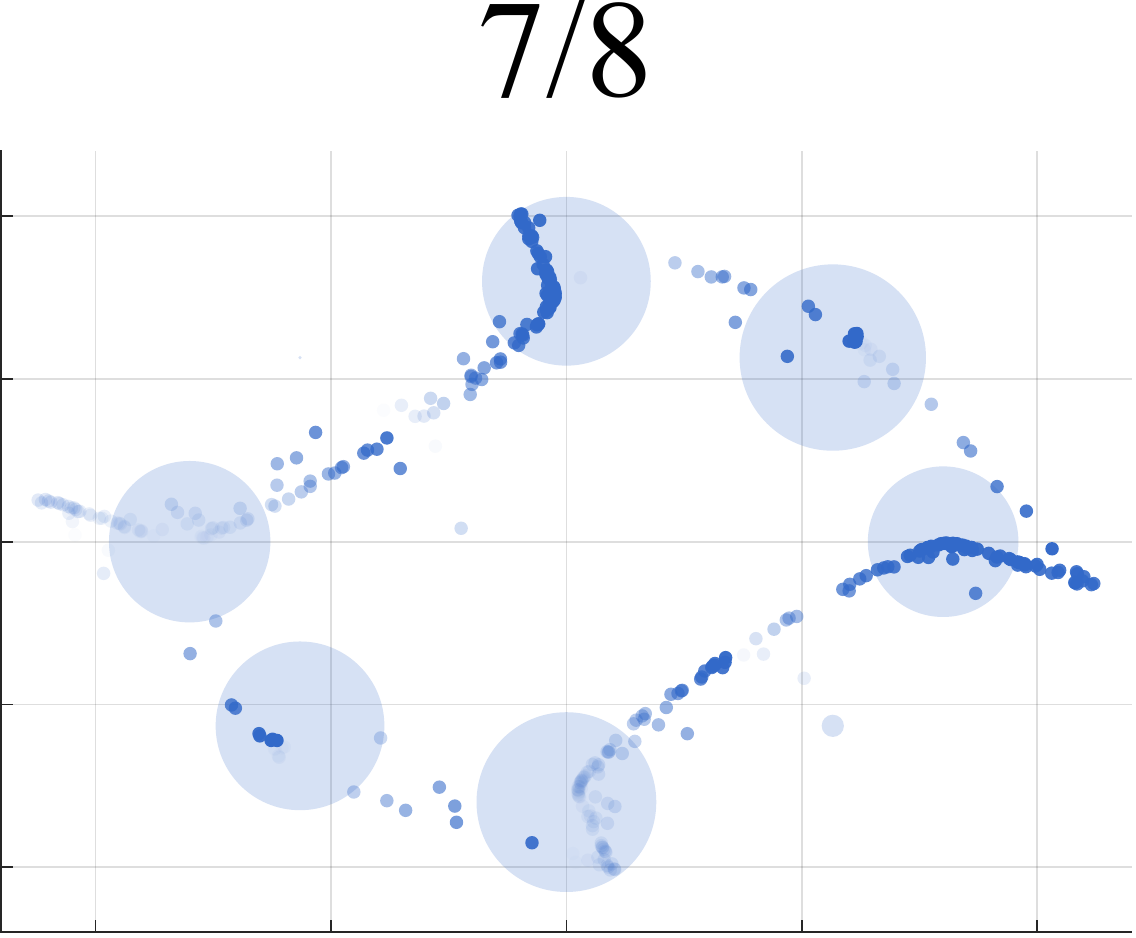} & \includegraphics[height=1.9cm]{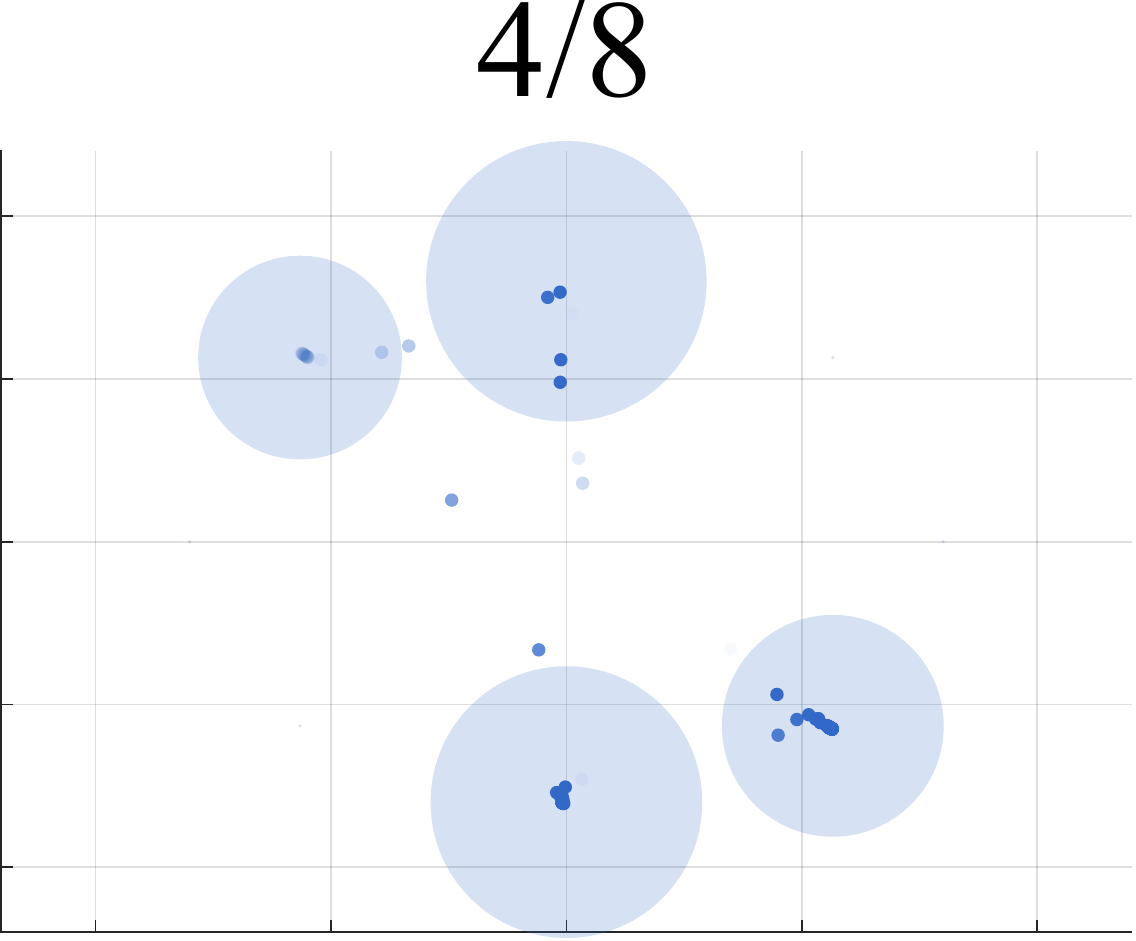}&& \multirow{7}{*}{\textbf{w/}}\\
			& \includegraphics[height=1.9cm]{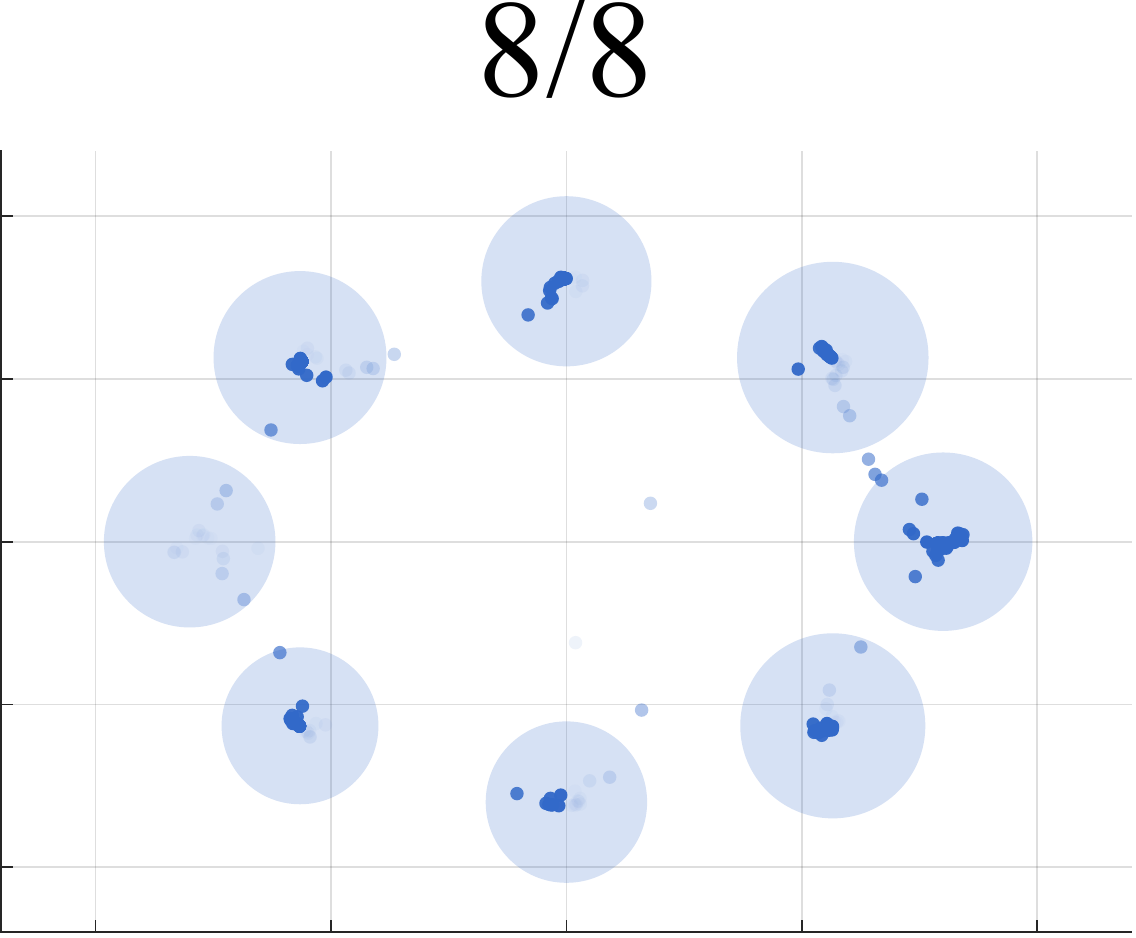} & \includegraphics[height=1.9cm]{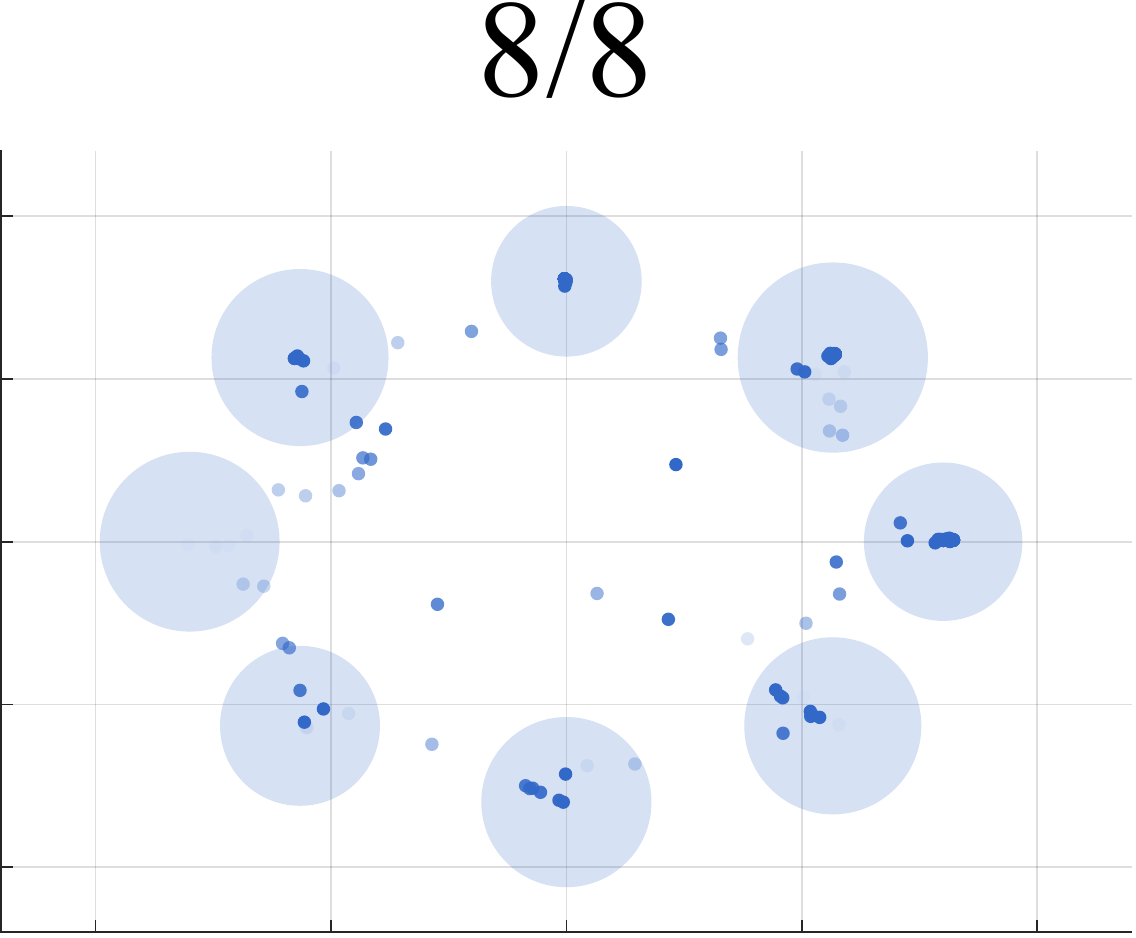} &\includegraphics[height=1.9cm]{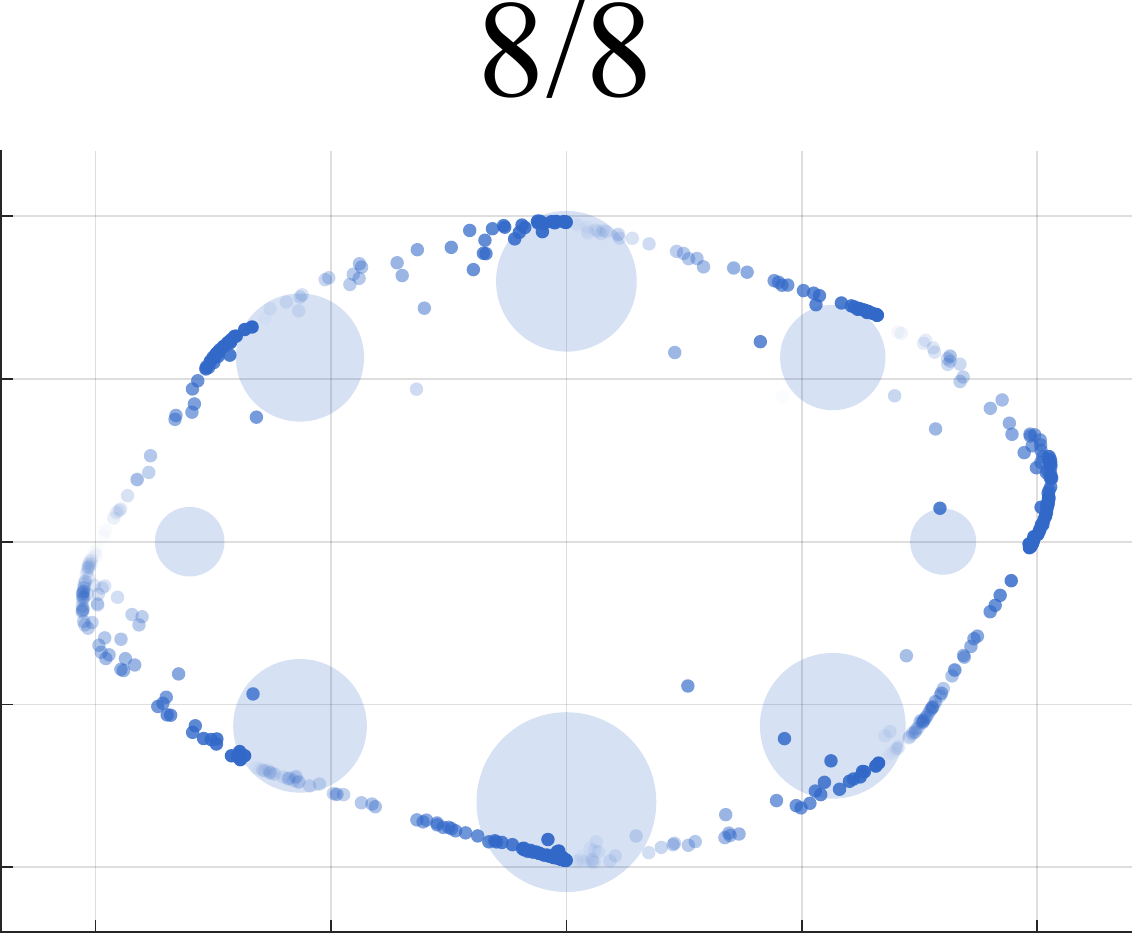} &  \includegraphics[height=1.9cm]{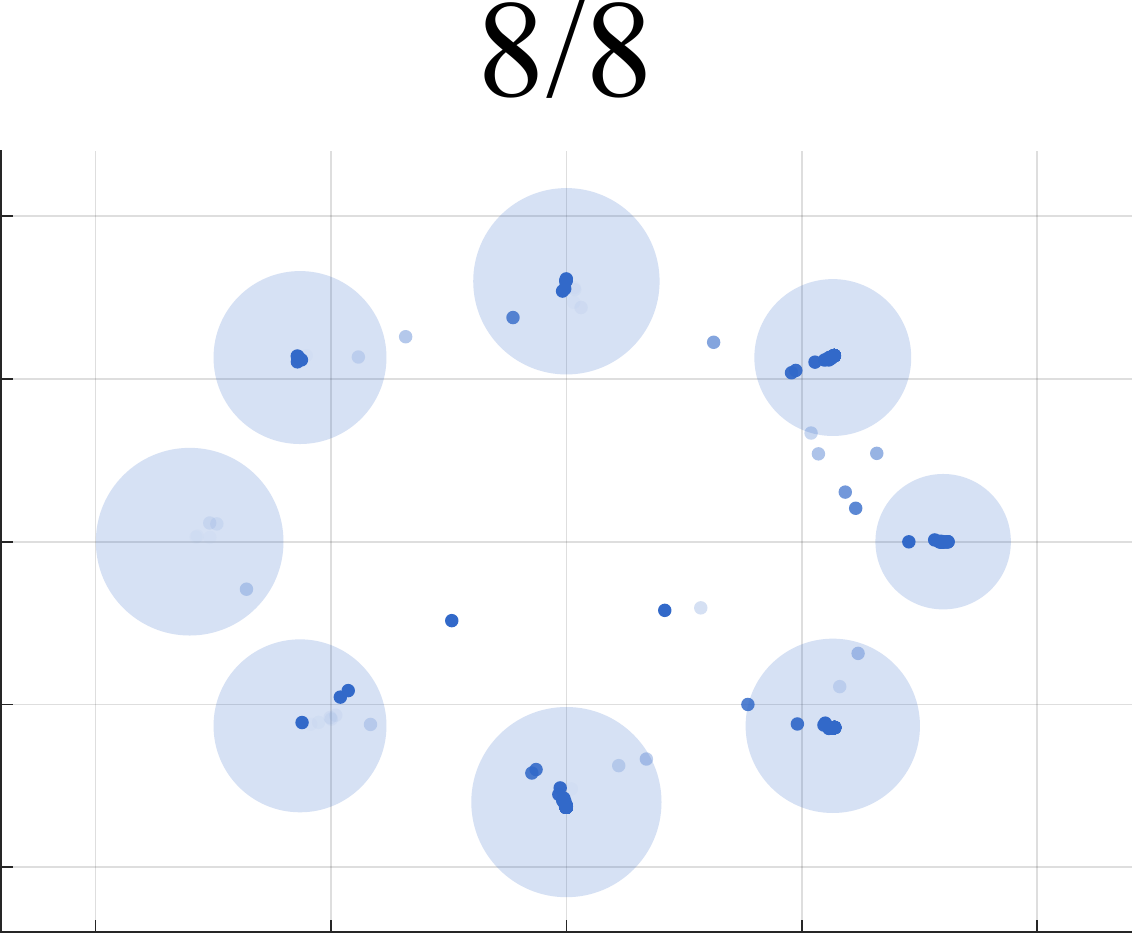} && \\
						\multicolumn{7}{c}{\textbf{2D Ring}} \\ \specialrule{0em}{1pt}{1pt} 
						\hline \specialrule{0em}{1pt}{1pt} 
						\vspace{0.1cm} 
					\multirow{8}{*}{\footnotesize \textbf{Target}} & \footnotesize \textbf{VGAN} & \footnotesize \textbf{WGAN} & \footnotesize \textbf{ProxGAN} & \footnotesize \textbf{LCGAN} & & \multirow{7}{*}{\textbf{w/o}} \\
					
					\vspace{-0.4cm}			%\multirow{2}{*}{\includegraphics[height=1.9cm]{fig/rand/target10div}}&\includegraphics[height=1.9cm]{fig/rand/GAN10div}& \includegraphics[height=1.9cm]{fig/rand/WGAN10div} & \includegraphics[height=1.9cm]{fig/rand/prox10div} & \includegraphics[height=1.9cm]{fig/rand/LC10div}& w/o\\
			%			& \includegraphics[height=1.9cm]{fig/rand/GN10div} & \includegraphics[height=1.9cm]{fig/rand/WGANGN10div} &\includegraphics[height=1.9cm]{fig/rand/proxGN10div} &  \includegraphics[height=1.9cm]{fig/rand/LCGN10div}& w/\\
			%			\multicolumn{6}{c}{ 2D Rand} \\ 
			%			\multirow{2}{*}{\includegraphics[height=1.9cm]{fig/grid/target10div}}&\includegraphics[height=1.9cm]{fig/grid/GAN10div}& \includegraphics[height=1.9cm]{fig/grid/WGAN10div} & \includegraphics[height=1.9cm]{fig/grid/prox10div} & \includegraphics[height=1.9cm]{fig/grid/LC10div}& w/o\\
			%			& \includegraphics[height=1.9cm]{fig/grid/GN10div} & \includegraphics[height=1.9cm]{fig/grid/WGANGN10div} &\includegraphics[height=1.9cm]{fig/grid/proxGN10div} &  \includegraphics[height=1.9cm]{fig/grid/LCGN10div} & w/ \\
			%			\multicolumn{6}{c}{ 2D Grid} \\ 
			\multirow{12}{*}{\includegraphics[height=1.9cm,trim=0 0 0 0,clip]{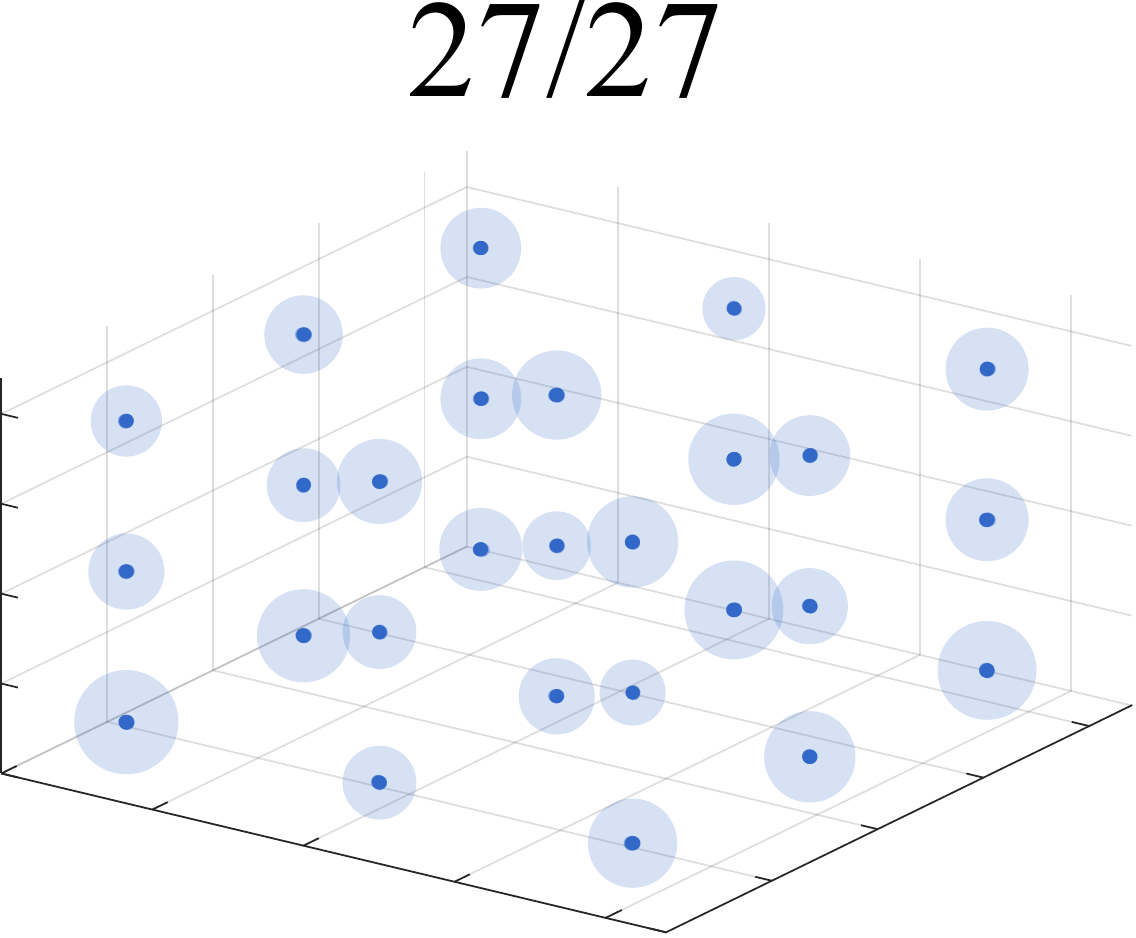}}& &&& & & \\	 
			&\includegraphics[height=1.9cm,trim=0 0 0 0,clip]{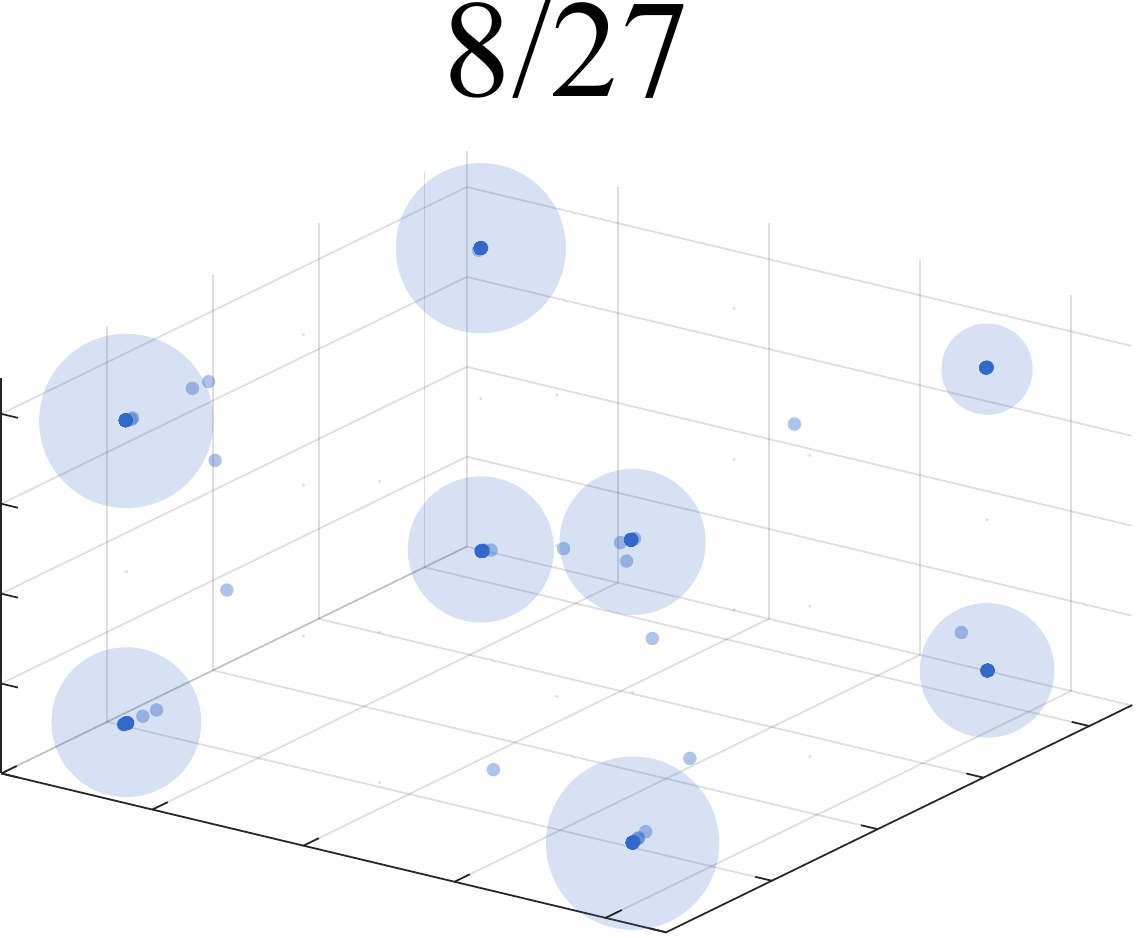}& \includegraphics[height=1.9cm,trim=0 0 0 0,clip]{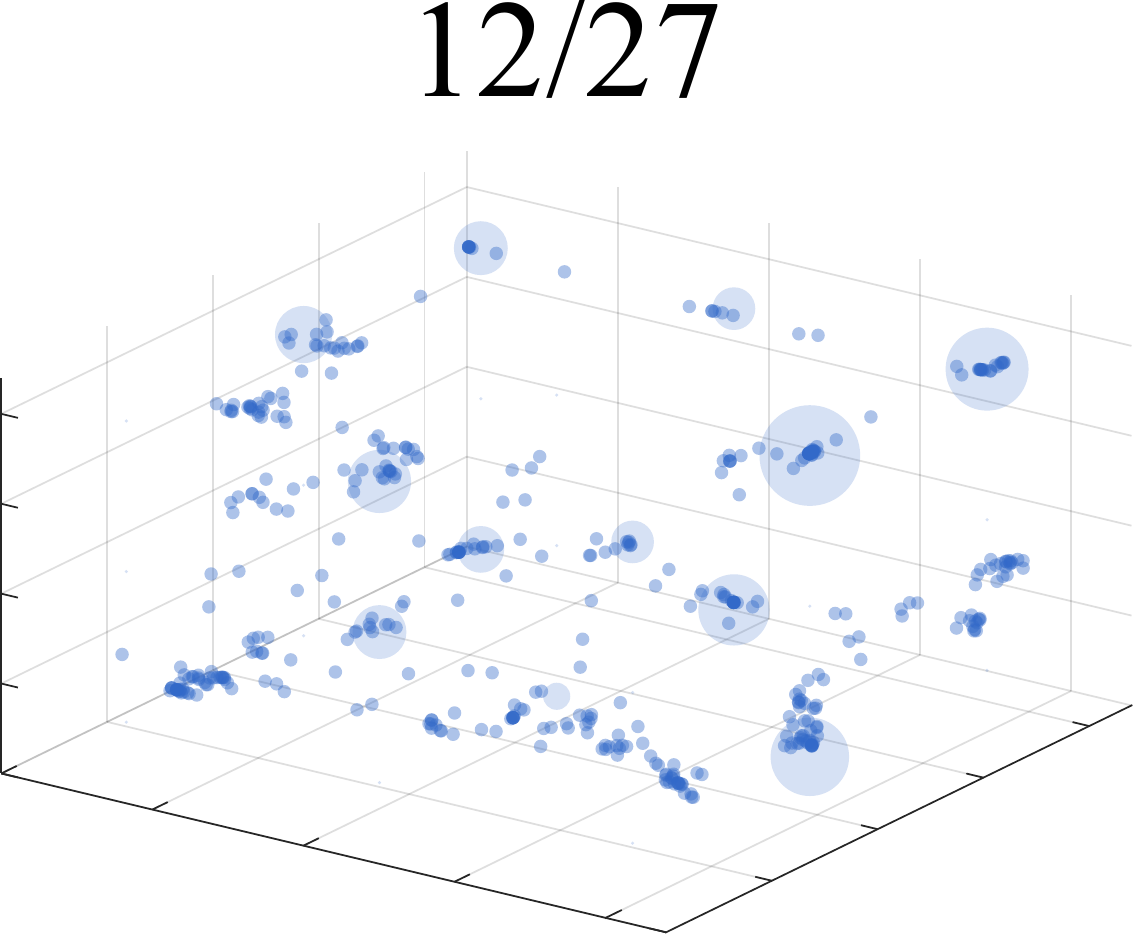} & \includegraphics[height=1.9cm,trim=0 0 0 0,clip]{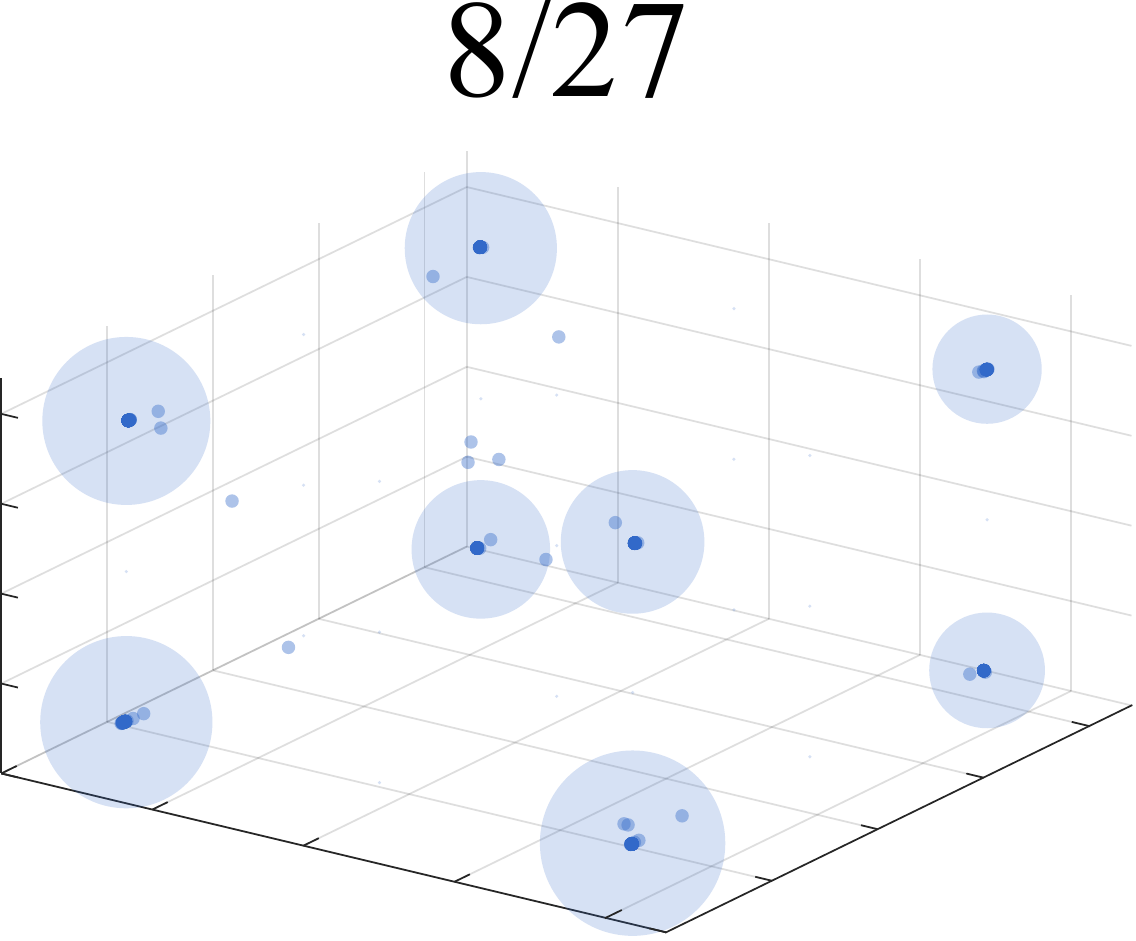} & \includegraphics[height=1.9cm,trim=0 0 0 0,clip]{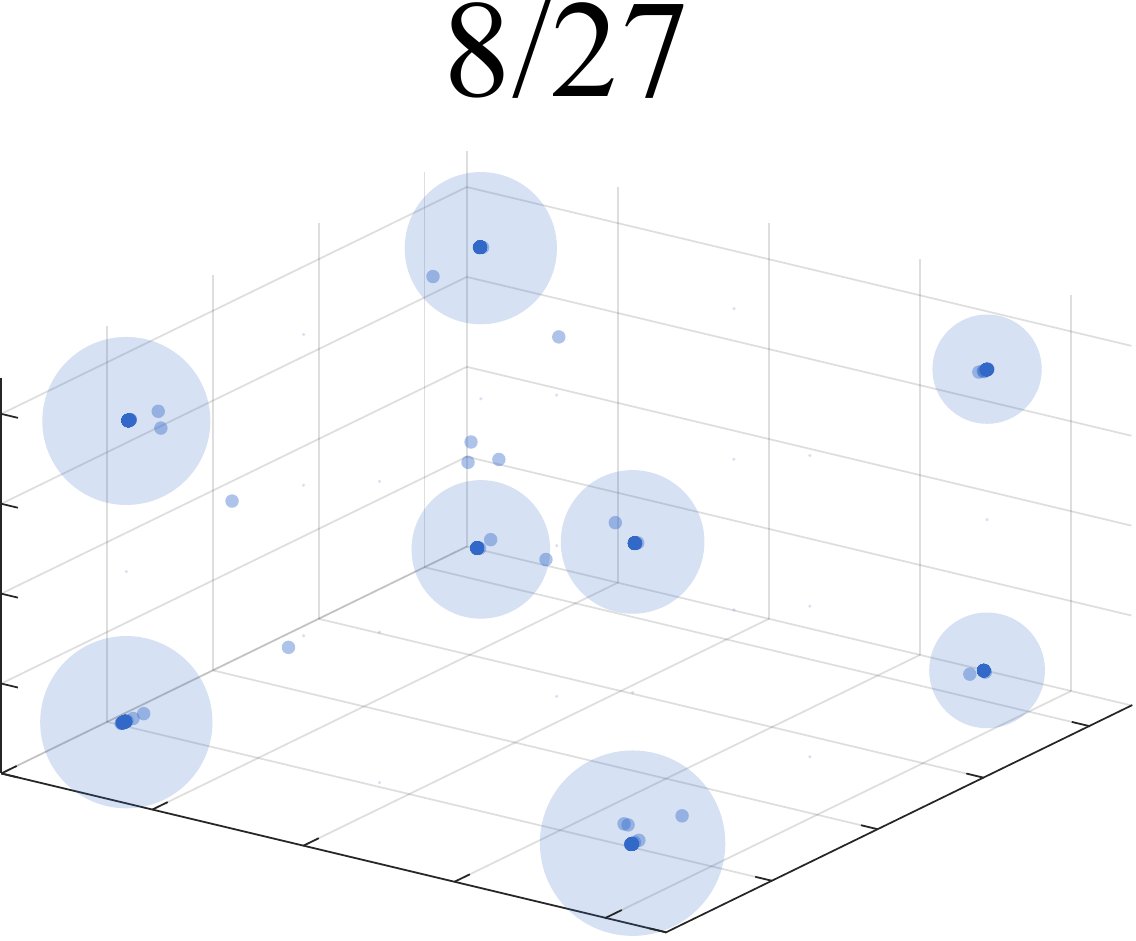} & & \multirow{6}{*}{\textbf{w/}}\\
			& \includegraphics[height=1.9cm,trim=0 0 0 0,clip]{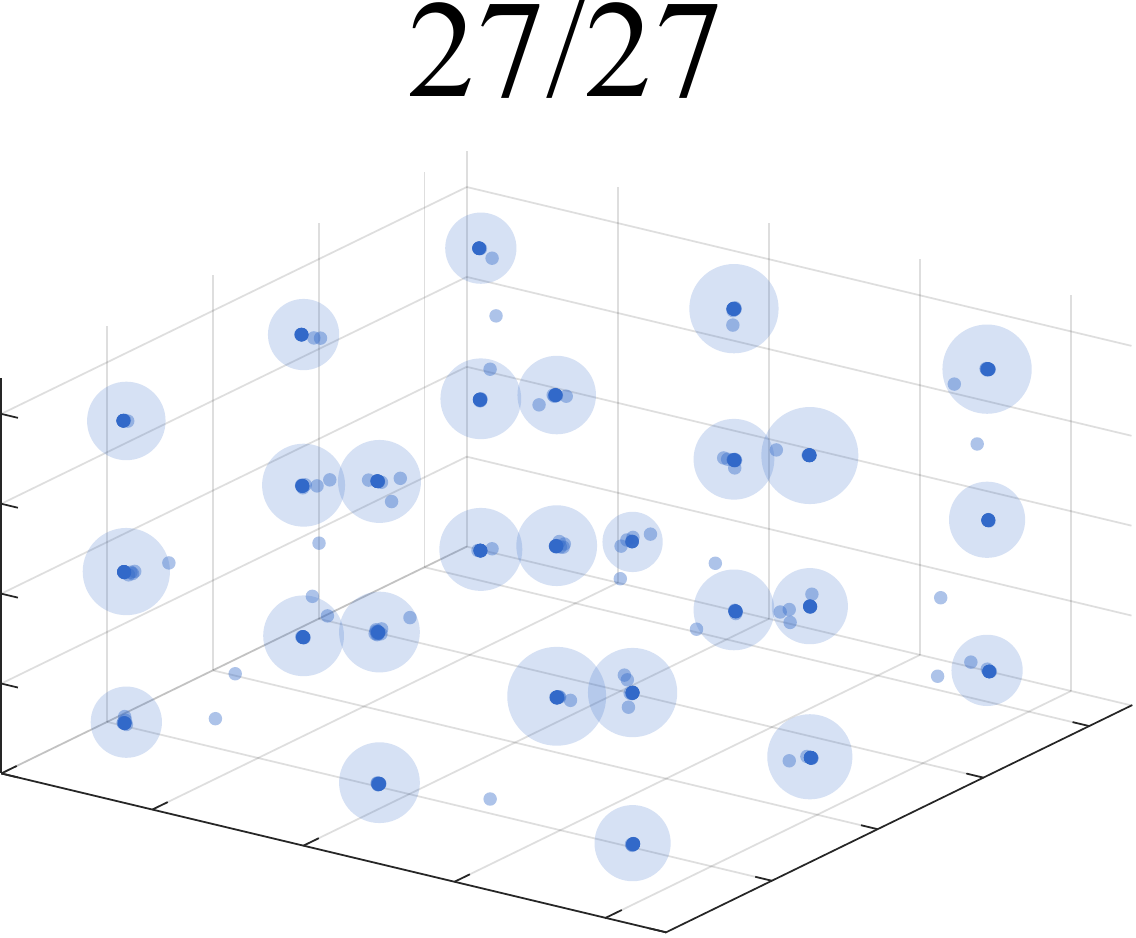} & \includegraphics[height=1.9cm,trim=0 0 0 0,clip]{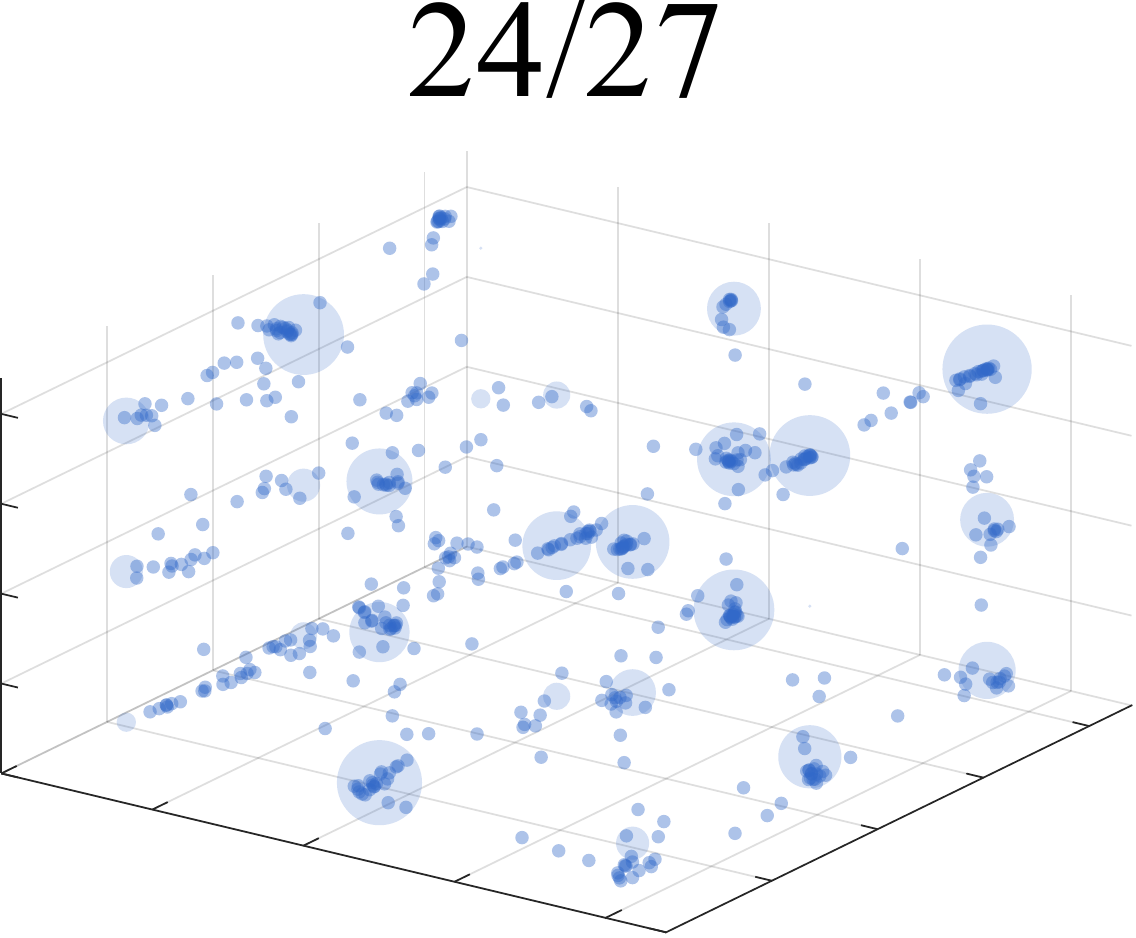} &\includegraphics[height=1.9cm,trim=0 0 0 0,clip]{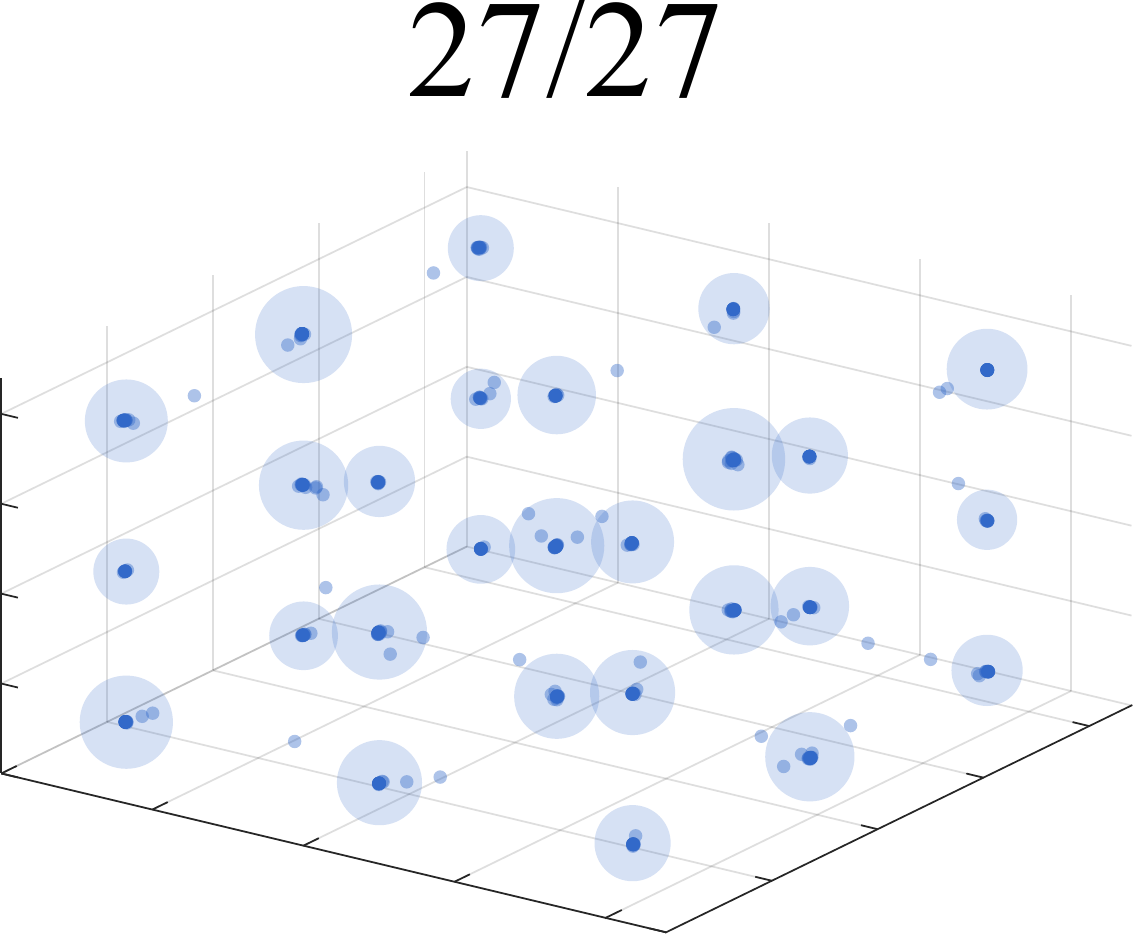} &  \includegraphics[height=1.9cm,trim=0 0 0 0,clip]{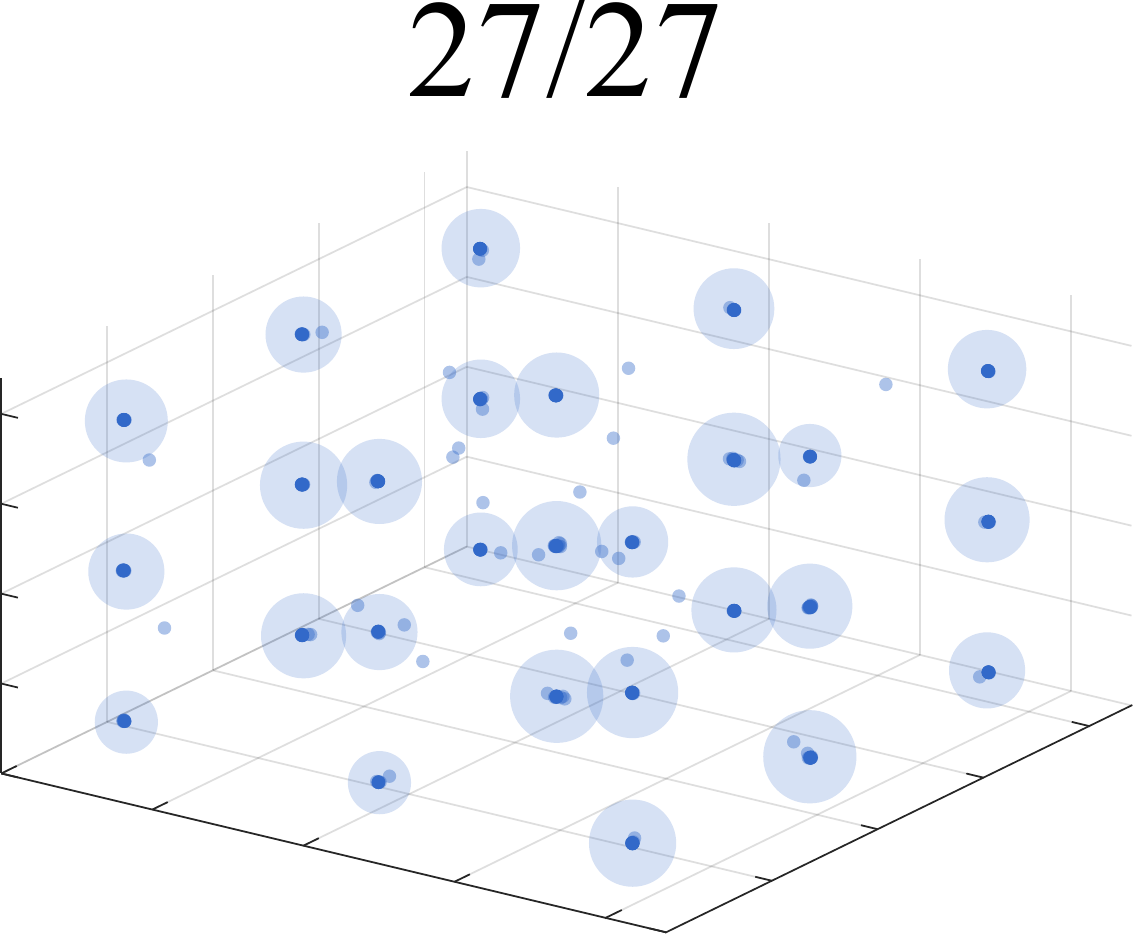} & & \\ 
			\multicolumn{7}{c}{\textbf{3D Cube}} \\ 
			\hline \toprule[0.65pt]
		\end{tabular}
	\end{center}
	\vspace{-0.2cm}
	\caption{Comparison results  among four mainstream GAN methods on two synthetic MOG distribution datasets. Heatmaps for all distributions are generated through the same iterative steps. The diversity of generated samples (number of generated classes/number of target classes) is listed at the top. All experiments were terminated when their performance stabilized. The shading of the dots represents the density of the final distribution, with darker dots representing greater density.}
	\label{fig:syn_mode}
\end{figure*}

To further verify the flexibility and effectiveness of our method, both visual and quantitative comparisons among various existing GAN methods (VGAN, LCGAN, WGAN and ProxGAN) on two synthetic datasets are conducted. Noteworthy, the same number of iterations among various GAN methods is strictly enforced to facilitate a fair comparison. 
Compared with original GAN methods, it can be observed in Table~\ref{tab:syn_metric} and Figure~\ref{fig:syn_mode} that GAN variants combined with our BRC framework exhibit better performance in the validation dataset. 

Table~\ref{tab:syn_metric} provides a comparison of results among existing GAN methods (i.e., VGAN, LCGAN, WGAN and ProxGAN) with or without our BRC framework across three metrics listed above (i.e., FID, JS and Mode) on two synthetic datasets (i.e., 2D Ring and 3D Cube). 
As can be seen, all of the results illustrate that GAN methods with our BRC framework can outperform original GAN methods for all metrics. In particular, the mode collapse problem has been well improved.
It can be observed that WGAN combined with BRC can obtain the smallest FID score in 2D ring distribution, and gets the lowest JS score in 3D cube dataset. 
Experiments on all synthesized data validate the superiority and flexibility for all existing GAN methods retrained in our proposed BRC framework.   
   
\begin{table*}[htbp]
	\vspace{-0.4cm}
	\begin{center}  \footnotesize
		\caption{Comparison of evaluation results of four mainstream GAN methods (i.e., VGAN, LCGAN, WGAN, and ProxGAN) with or without  BRC. We report the average FID$\downarrow$, JSD $\downarrow$ and Mode $\uparrow$ scores on two MOG synthesized datasets. Best and second best results are \textbf{highlighted} and \underline{underlined}.}
		\vspace{0.1cm}
		\label{tab:syn_metric}%
		\setlength{\tabcolsep}{1.5mm}{
			\begin{tabular}{|c|c|c|c|c|c|c|c|}
				\hline
				\multirow{2}[0]{*}{Method} & \multirow{2}[0]{*}{BRC} &\multicolumn{3}{c|}{2D Ring (Max Mode=8)}&	\multicolumn{3}{c|}{3D Cube (Max Mode=27)}\\
				\cline{3-5}
				\cline{6-8}		
				& & FID$\downarrow$   & JS$\downarrow$   & Mode$\uparrow$    & \footnotesize FID$\downarrow$   &  JS$\downarrow$  & Mode$\uparrow$\\		
				\hline
				\hline
				\multirow{2}[0]{*}{\footnotesize VGAN} & \xmark & 193.20$\pm$65.30 & 0.63$\pm$0.12 & 3.50$\pm$1.00&  12.28$\pm$26.10 & \underline{0.48$\pm$0.08} & 8.50$\pm$1.00\\
				\cline{2-8}
				& \cmark & 34.16$\pm$10.23 & \underline{0.40$\pm$0.16} & \textbf{7.50$\pm$1.00}& \textbf{0.52$\pm$0.16} & 0.61$\pm$0.30  & \underline{23.00$\pm$1.40}  \\
				\hline
				\multirow{2}[0]{*}{\footnotesize LCGAN} & \xmark & 15.01$\pm$15.30 & 0.63$\pm$0.12 & 3.50$\pm$1.00& 68.80$\pm$60.85&0.87$\pm$0.17&6.00$\pm$3.16\\
				\cline{2-8}
				& \cmark & 1.19$\pm$1.23 & \underline{0.40$\pm$0.16} & \textbf{7.50$\pm$1.00}&30.5$\pm$45.81 & 0.70$\pm$0.30 & 17.33$\pm$14.15 \\
				\hline
				\multirow{2}[0]{*}{WGAN} & \xmark  & \underline{0.42$\pm$0.21} & 0.65$\pm$0.13 & 6.25$\pm$1.53 & 12.00$\pm$0.90 & 0.62$\pm$0.07 & 15.00$\pm$0.41 \\
				\cline{2-8}
				& \cmark & \textbf{0.16$\pm$0.09} & \textbf{0.27$\pm$0.25} & \underline{7.00$\pm$0.82} & \underline{0.80$\pm$0.20} & \textbf{0.21$\pm$0.03}   & \textbf{24.00$\pm$0.51}  \\
				\hline
				\multirow{2}[0]{*}{ProxGAN} & \xmark & 15.01$\pm$15.30 & 0.63$\pm$0.12 & 3.50$\pm$1.00&111.64$\pm$63.96 & 0.85$\pm$0.03 & 3.67$\pm$2.89 \\
				\cline{2-8}
				& \cmark & 8.82$\pm$17.25 & 0.58$\pm$0.26 & 6.50$\pm$3.00 & 52.10$\pm$86.98 & 0.69$\pm$0.30 & 17.67$\pm$14.47\\			
				\hline
			\end{tabular}%
		}
	\end{center}
\vspace{-0.4cm}	
\end{table*}% 

Figure~\ref{fig:syn_mode} shows a comparison of results regarding the number of samples generated among current advanced GAN methods with or without our BRC framework. VGAN and LCGAN in our framework already generate all modes and are significantly better than corresponding original GAN methods in terms of similarity to the real target under 2D ring distribution. 
It can be seen original GANs only capture a small number of distributions, getting into the most severe mode collapse dilemma and failing to achieve satisfactory performance. While under our framework, these method can capture a relatively large number of distributions with the assistance of our method, and generate abundant real distributions. 
On 3D Cube distribution, ProxGAN and LCGAN combined with our method can fit almost all Gaussian distributions well with plenty of details. Experiments on all synthesized data validate the superiority of original GAN methods trained in our framework.  

\begin{wrapfigure}{r}{9.7cm}
	\vspace{-0.6cm}
	\begin{minipage}[t]{1.0\linewidth}
		\centering 
		\makeatletter\def\@captype{table}\makeatother
		\caption{Comparison of the time complexity (S) and space complexity (MB) on 2D and 3D MOG datasets for each step of calculating the gradient of $\bm{\theta}_{G}$ against existing gradient-based algorithms. Note that ``T'' and ``M'' represent ``Time'' and ``Memory'', respectively. Best and second best results are \textbf{highlighted} and \underline{underlined}.
		} \vspace{0.1cm} 
		\label{table:Tnet} 
		\setlength{\tabcolsep}{1.85mm}{
			\begin{tabular}{|c|c|c|c|c|c|c|}				
				\hline
				\footnotesize Dataset  & \footnotesize Metric & \footnotesize RHG   & \footnotesize BDA   & \footnotesize CG    & \footnotesize Neumann & \footnotesize IGA \\
				\hline   
				\hline
				\multirow{2}{*}{\footnotesize 2D MOG}&\footnotesize T &\footnotesize 1083.97&\footnotesize 1512.74 &\footnotesize 317.96 & \underline{\footnotesize 311.16}&\textbf{\footnotesize 107.77} \\
				\cline{2-7} 
				& \footnotesize M  &\footnotesize 94.85&\footnotesize 106.31& \footnotesize 45.56&  \underline{\footnotesize 43.92}&\textbf{\footnotesize 16.34} \\
				\hline 
				\multirow{2}{*}{\footnotesize 3D MOG}&  \footnotesize T &\footnotesize 1031.49& \footnotesize 1537.86 & \footnotesize 373.62 & \underline{\footnotesize 309.21}&\textbf{\footnotesize 94.98} \\
				\cline{2-7}
				& \footnotesize M  &\footnotesize 74.90&\footnotesize 96.85& \footnotesize 23.16&  \underline{\footnotesize 21.05}&\textbf{\footnotesize 12.56} \\
				\hline 
			\end{tabular}\label{tab:time2}}
	\end{minipage}
	\vspace{-0.2cm}
\end{wrapfigure}
\textbf{Computational efficiency.} We also compare time and space complexity among our strategy and current mainstream hierarchical optimization solution strategies on 2D and 3D MOG synthetic datasets.
As can be seen from the Table~\ref{tab:time2}, the implicit gradient methods (i.e. CG~\cite{pedregosa2016hyperparameter} and Neumann~\cite{lorraine2020optimizing}) keep the computational complexity lower compared to the explicit gradient methods (i.e. RHG~\cite{franceschi2017forward} and BDA~\cite{liu2022general}) because they are efficient to avoid the computationally expensive Hessian inverse. Our method outperforms these methods and can guarantee less time consumption and less memory consumption, especially in high-dimensional tasks.  
 \vspace{-0.2cm}
\subsection{Validation on Real-world Image Generation Task} 
\vspace{-0.2cm}
\textbf{Implementation Details.} In this part, we evaluate the flexibility and effectiveness of our framework on the real-world image generation task.  
In terms of image generation, we first use the popular CIFAR10~\cite{krizhevsky2010convolutional},  CIFAR100~\cite{krizhevsky2009learning} benchmark for evaluation and in-depth analysis of our method training strategy. 
Both CIFAR10 and CIFAR100 are composed of 60,000 colour images with $32\times32$ size, where 50,000 images are used as training set and the rest 10,000 images are used as test set. 
%The CIFAR10 dataset contains 10 classes, and the CIFAR100 contains 100 classes. 
We use Inception Score (IS)~\cite{salimans2016improved} for evaluating generation quality and diversity, and FID for capturing model issues, e.g., mode collapse.
All the results show our algorithm can further facilitate the training of original GAN methods and obtain better sample generation quality and performance score.  We train  DCGAN,  LSGAN and SNGAN by executing 200k, 200k and 20k  iterations, respectively.  For the network architecture and other training details, we follow the settings in \cite{kang2021ReACGAN}.

%Further, we explore the effectiveness of our method compared to current advanced GANs in real-world vision applications, and the analysis will focus on several cutting-edge mainstream low-level, high-level vision and reinforcement learning applications, 
\begin{wrapfigure}{r}{7.5cm}
	\vspace{-0.6cm} 
	\begin{minipage}[t]{1.0\linewidth}
		\centering 
		\makeatletter\def\@captype{table}\makeatother
		\caption{ Comparison results of FID and IS score on CIFAR10 and CIFAR100 dataset training with or without BRC. Best and second best results are \textbf{highlighted} and \underline{underlined}. } \vspace{0.1cm} 
		\label{tab:real_is} 
		
		\setlength{\tabcolsep}{2.4mm}{
			\begin{tabular}{|c|c|c|c|c|c|}
				\hline  
				 \multirow{2}[0]{*}{\footnotesize Method} & \multirow{2}[0]{*}{\footnotesize BRC}& \multicolumn{2}{c|}{\footnotesize CIFAR10}&  \multicolumn{2}{c|}{\footnotesize CIFAR100}\\ 
				\cline{3-6}   		
				& &\footnotesize IS$\uparrow$&\footnotesize FID$\downarrow$ &\footnotesize IS$\uparrow$&\footnotesize FID$\downarrow$\\
				\hline 
				\hline
				\footnotesize \multirow{2}[0]{*}{\footnotesize DCGAN} & \xmark & \footnotesize 6.63& \footnotesize 49.03 & \footnotesize 6.56& \footnotesize 57.37    \\
				\cline{2-6}  
				&\cmark &\footnotesize7.06& \footnotesize 42.23 & \footnotesize 6.87  & \footnotesize 44.18    \\
				\hline 
				\multirow{2}[0]{*}{\footnotesize LSGAN} & \xmark &\footnotesize 5.57& \footnotesize 66.68 & \footnotesize 3.81& \footnotesize 145.54 \\
				\cline{2-6}  
				&\cmark &\underline{\footnotesize 7.54}&\footnotesize 32.50 & \footnotesize 7.30 & \footnotesize 35.72 \\
				\hline 
				\multirow{2}[0]{*}{\footnotesize SNGAN} &\xmark & \footnotesize 7.48&\underline{\footnotesize 26.51} &\underline{\footnotesize 7.99}&\underline{\footnotesize 25.33}    \\ 
				\cline{2-6}  			
				&\cmark  &\textbf{\footnotesize 7.58}&\textbf{\footnotesize 22.81}& \textbf{\footnotesize 8.27}&\textbf{\footnotesize 21.37}   \\
				\hline 
			\end{tabular}% 
	}
\end{minipage}	
	\vspace{-0.3cm}
\end{wrapfigure}

\begin{figure}[htbp]
	\begin{center}
		\begin{tabular}{c@{\extracolsep{0.8em}}c@{\extracolsep{0.8em}}c@{\extracolsep{0.8em}}c@{\extracolsep{0.8em}}c@{\extracolsep{0.8em}}}
			\includegraphics[height=2.8cm,trim=0 0 0 0,clip]{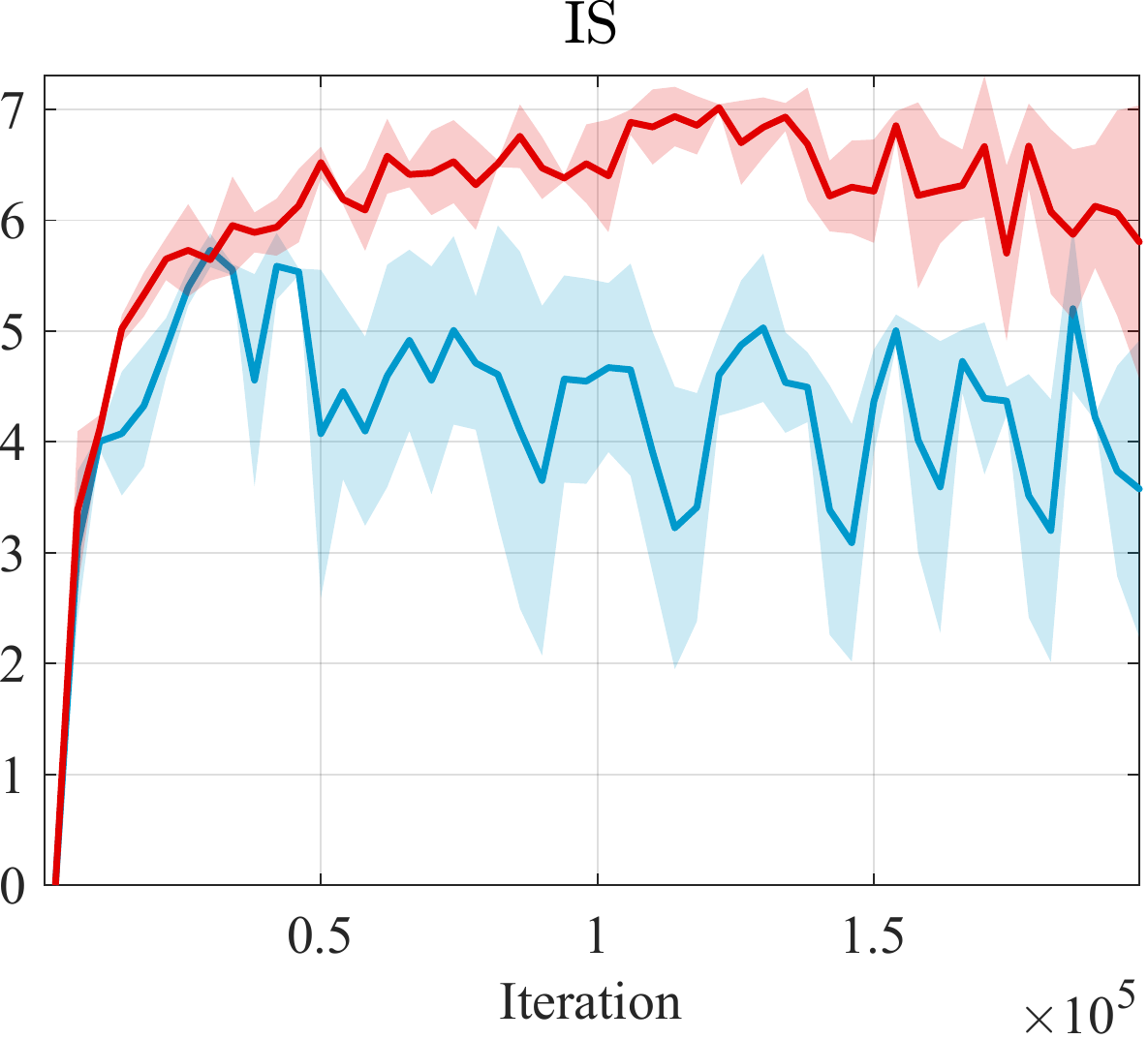} &\includegraphics[height=2.8cm,trim=0 0 0 0,clip]{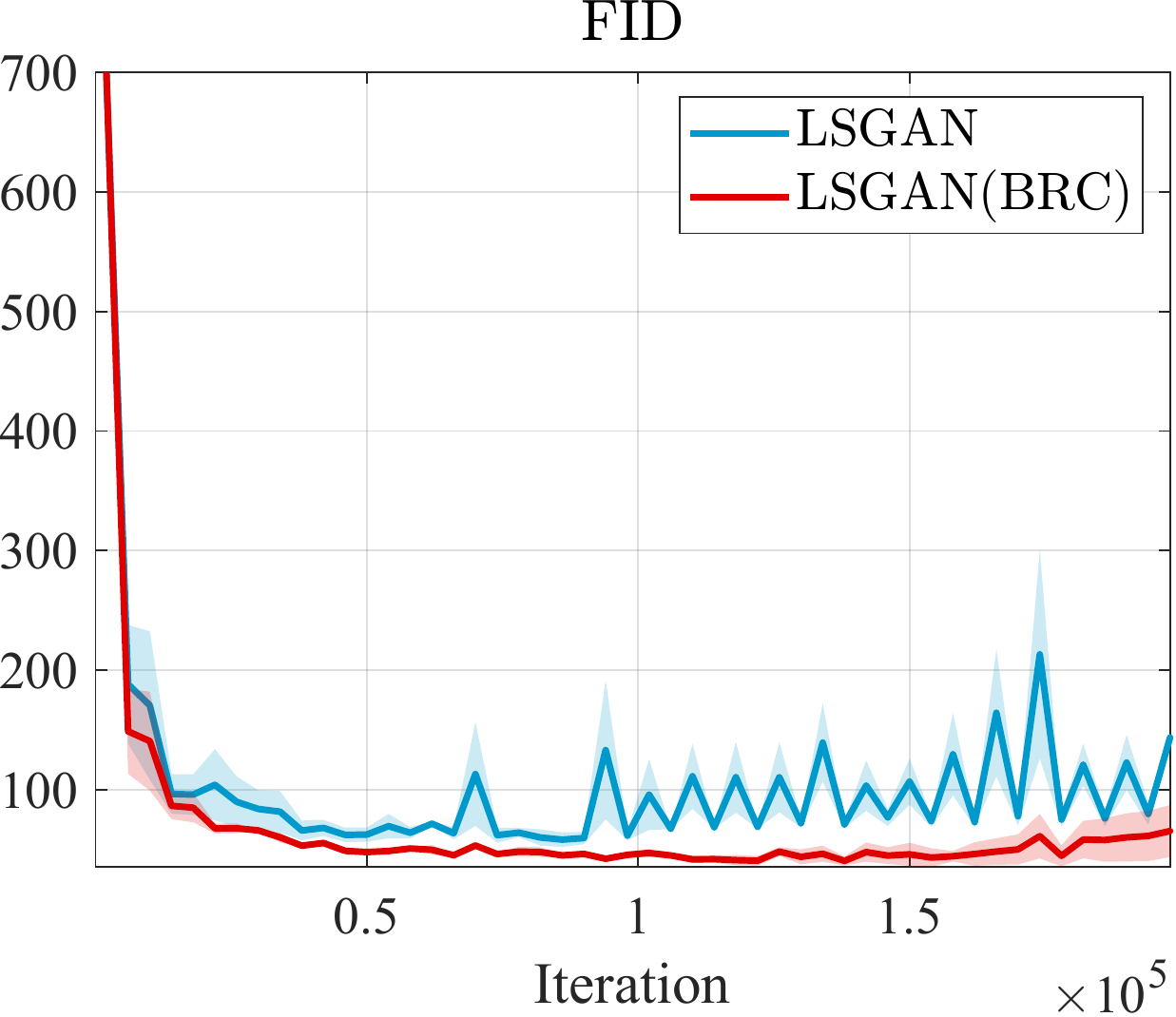} && \includegraphics[height=2.8cm,trim=0 0 0 0,clip]{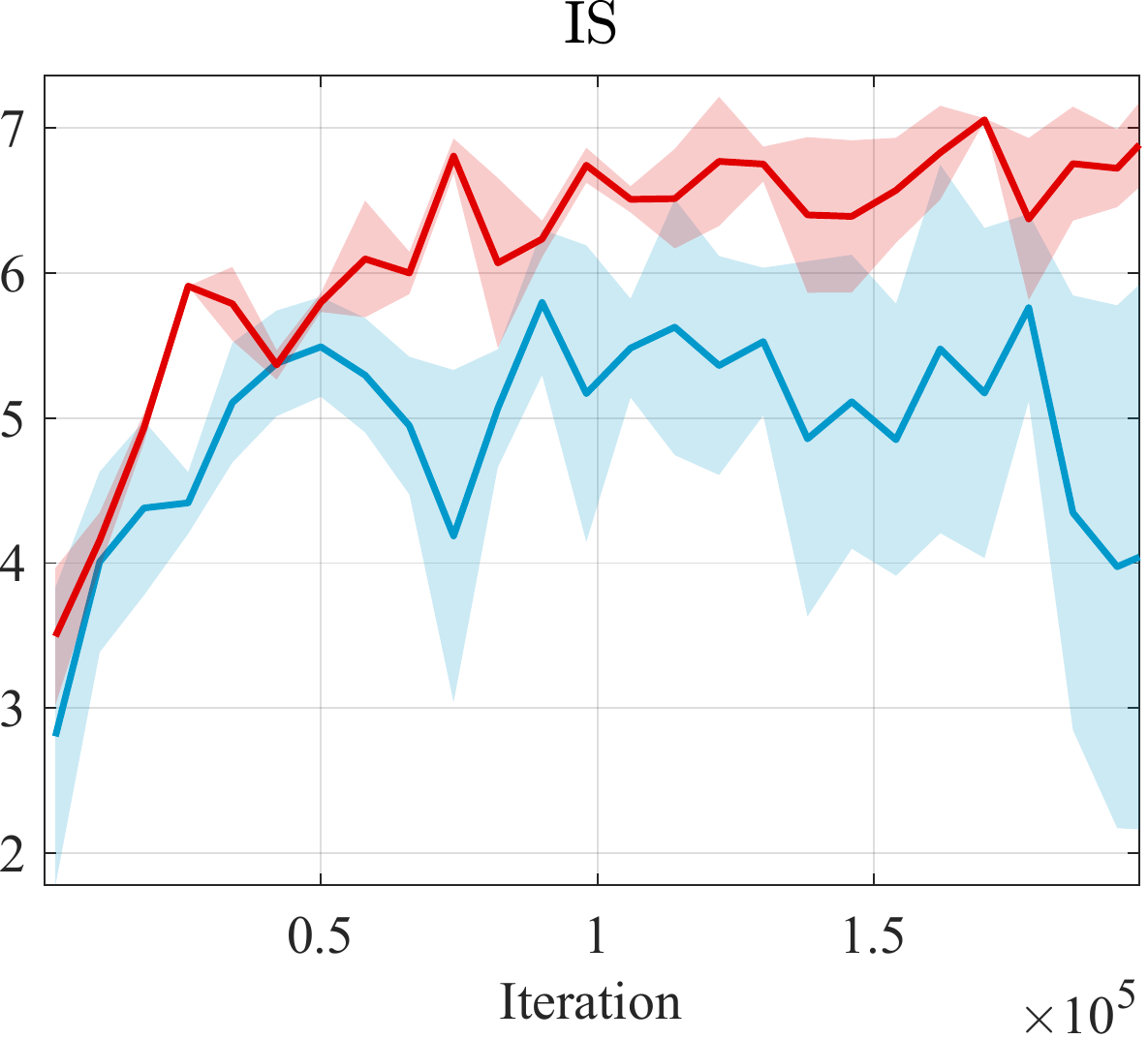} &\includegraphics[height=2.8cm,trim=0 0 0 0,clip]{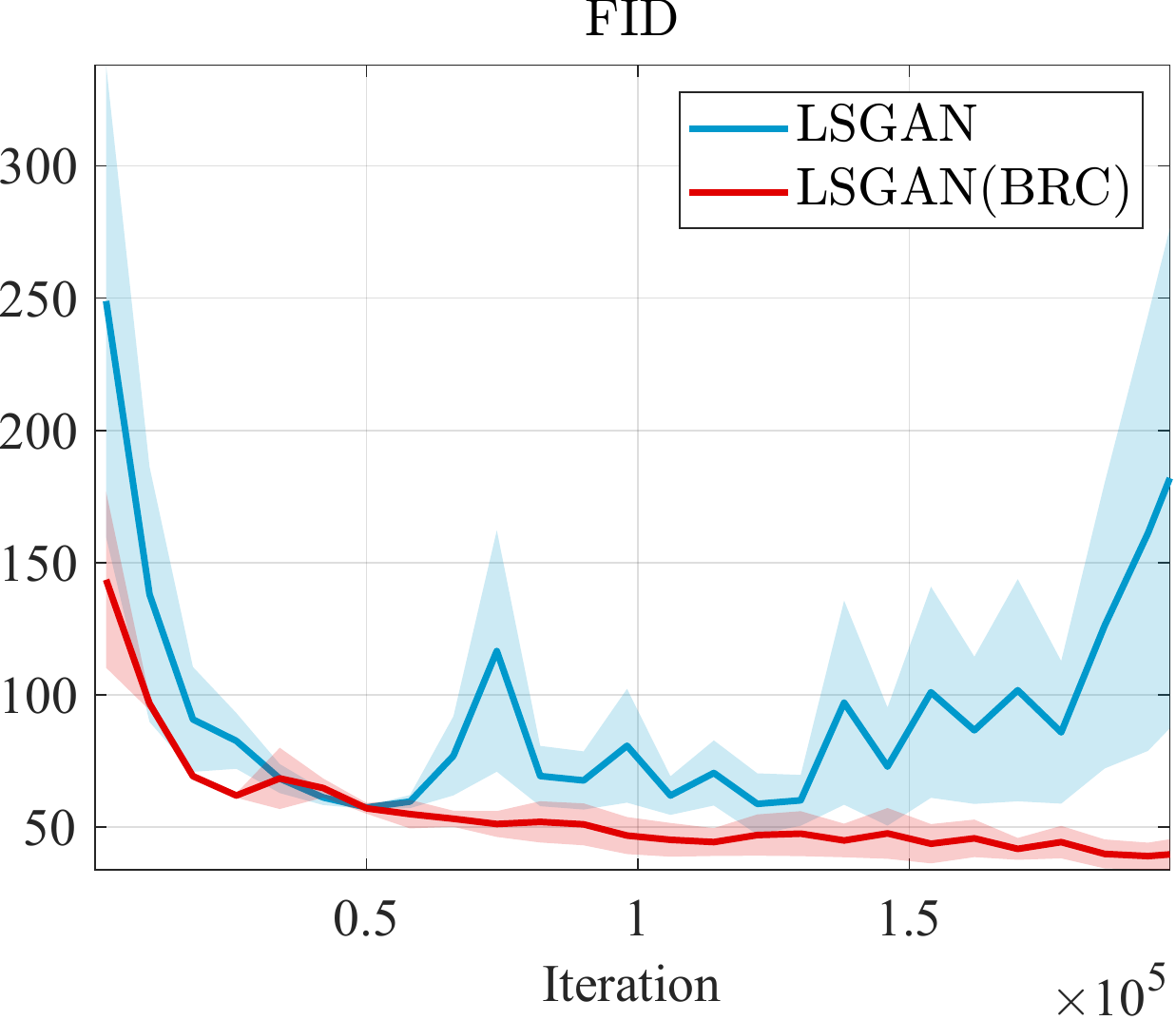}\\
			\multicolumn{2}{c}{\footnotesize CIFAR10 } && 
			\multicolumn{2}{c}{\footnotesize CIFAR100}\\
		\end{tabular}
	\end{center}
	\vspace{-0.2cm}
	\caption{Comparison of training efficiency results measured by FID and IS for LSGAN networks on CIFAR10 dataset (listed left) and CIFAR100 datasets (listed right). More stable training convergence curves and better FID and IS performance scores can be obtained by LSGAN combined with BRC.}
	\label{fig:real_is}
\end{figure}

%
%\begin{figure*}[h!]
%	\begin{center}
%		\begin{tabular}{c@{\extracolsep{0.1em}}c@{\extracolsep{0.1em}}}
%			\includegraphics[width=0.48\linewidth,trim=0 0 0 0,clip]{fig/stylegan/1}&\includegraphics[width=0.48\linewidth,trim=0 0 0 0,clip]{fig/stylegan/2}\\\specialrule{0em}{-2pt}{-3.5pt}			
%			\multicolumn{2}{c}{\footnotesize StyleGAN} \\ 
%			\includegraphics[width=0.48\linewidth,trim=0 0 0 0,clip]{fig/stylegan/3}&\includegraphics[width=0.48\linewidth,trim=0 0 0 0,clip]{fig/stylegan/4}\\\specialrule{0em}{-2pt}{-3.5pt}	
%			\multicolumn{2}{c}{\footnotesize StyleGAN(BRC)}\\
%		\end{tabular}
%	\end{center}
%	\vspace{-0.4cm}
%	\caption{Comparison results of face generation on CelebA-HQ dataset~~\cite{liu2015deep} regarding whether the StyleGAN network is retrained within our framework. All experiments are terminated when their performance reaches plateau. 	It can be seen that StyleGAN trained in our framework is capable of generating more abundant face details. }
%	\label{fig:real_face}
%\end{figure*}

\textbf{Performance Evaluation.} First, we qualitatively evaluate three representative GAN networks (i.e., DCGAN~\cite{gao2018deep}, LSGAN~\cite{mao2017least}, SNGAN~\cite{miyato2018spectral})  combined with our BRC framework on CIFAR10 and CIFAR100 datasets. When not training in our BRC framework, the generated images through original GANs tend to appear as collapsed and distorted fake shapes. In contrast, retraining by our algorithm will enable the generated samples to contain more variety and abundant details.
As shown in Figure~\ref{fig:real_is}, the trend of two curves implies that LSGAN retrained in our framework is able to converge stably and obtain lower FID and higher IS scores.  
%It also demonstrates that our method achieves higher performance than the original training strategy, and sometimes far better. Especially on CIFAR10, LSGAN within our framework outperforms original LSGAN by a large margin, significantly reducing the performance gap between DCGAN and SNGAN in terms of IS and FID scores. 
Table~\ref{tab:real_is} also demonstrates that our method achieves higher performance than the original training strategy. Especially on CIFAR10, LSGAN within our framework outperforms original LSGAN by a large margin, significantly reducing the performance gap between DCGAN and SNGAN in terms of IS and FID scores.

%Based on StyleGAN as the backbone architecture, we also conduct face generation experiment on high-resolution CelebA dataset~\cite{liu2015deep}. CelebA is a large scale face attributes dataset, which consists of more than 200K celebrity images. Figure~\ref{fig:real_face} indicates that StyleGAN within our framework, denoted as StyleGAN (BRC), achieves the best overall style-content trade-off, which verifies the universality of our flexible solution strategy. Our method is in effective of producing real face structures and suppresses twist distortion. For other training details, we follow the same settings as StyleGAN. 
\vspace{-0.3cm} 
\section{Conclusions and Future Works}
\vspace{-0.2cm}  
To alleviate a series of issues like mode collapse, vanishing gradients and oscillations caused by the single-level minimax type GANs, this paper introduces a novel BRC framework based on a hierarchical optimization perspective to explicitly formulate the potential dependency of the generator on the discriminator. As a general learning framework, most of current mainstream GANs can reformulate the optimization of discriminator as a constraint and incorporate it into the generator optimization. On this basis, we also design a fast solution strategy that combines implicit function theory with an outer-product-based Hessian approximation technique, to mitigate the high complexity of original hierarchical strategies. Extensive quantitative and qualitative experimental results verify the effectiveness, flexibility and stability in both synthetic and real-world scenarios. Future research should be extended to apply BRC framework to more challenging and complex high-level visual fields, or to explore faster and more stable learning strategies while maintaining performance.

%\section*{Acknowledgment}
%This work is partially supported by the National Key R\&D Program of China (2020YFB1313503), the National Natural Science Foundation of China (Nos. 61922019, 61733002, and 61672125), LiaoNing Revitalization Talents Program (XLYC1807088), and the Fundamental Research Funds for the Central Universities.

\bibliographystyle{unsrtnat}
\bibliography{nips2022}

\begin{thebibliography}{31}
\providecommand{\natexlab}[1]{#1}
\providecommand{\url}[1]{\texttt{#1}}
\expandafter\ifx\csname urlstyle\endcsname\relax
  \providecommand{\doi}[1]{doi: #1}\else
  \providecommand{\doi}{doi: \begingroup \urlstyle{rm}\Url}\fi

\bibitem[Goodfellow et~al.(2014)Goodfellow, Pouget-Abadie, Mirza, Xu,
  Warde-Farley, Ozair, Courville, and Bengio]{goodfellow2014generative}
Ian Goodfellow, Jean Pouget-Abadie, Mehdi Mirza, Bing Xu, David Warde-Farley,
  Sherjil Ozair, Aaron Courville, and Yoshua Bengio.
\newblock Generative adversarial nets.
\newblock \emph{NeurIPS}, 27, 2014.

\bibitem[Li et~al.(2019)Li, Qi, Lukasiewicz, and Torr]{li2019controllable}
Bowen Li, Xiaojuan Qi, Thomas Lukasiewicz, and Philip~HS Torr.
\newblock Controllable text-to-image generation.
\newblock \emph{arXiv:1909.07083}, 2019.

\bibitem[Zhu et~al.(2017)Zhu, Park, Isola, and Efros]{zhu2017unpaired}
Jun-Yan Zhu, Taesung Park, Phillip Isola, and Alexei~A Efros.
\newblock Unpaired image-to-image translation using cycle-consistent
  adversarial networks.
\newblock In \emph{ICCV}, pages 2223--2232, 2017.

\bibitem[Ledig et~al.(2017)Ledig, Theis, Husz{\'a}r, Caballero, Cunningham,
  Acosta, Aitken, Tejani, Totz, Wang, et~al.]{ledig2017photo}
Christian Ledig, Lucas Theis, Ferenc Husz{\'a}r, Jose Caballero, Andrew
  Cunningham, Alejandro Acosta, Andrew Aitken, Alykhan Tejani, Johannes Totz,
  Zehan Wang, et~al.
\newblock Photo-realistic single image super-resolution using a generative
  adversarial network.
\newblock In \emph{CVPR}, 2017.

\bibitem[Wang et~al.(2018)Wang, Yu, Wu, Gu, Liu, Dong, Qiao, and
  Change~Loy]{wang2018esrgan}
Xintao Wang, Ke~Yu, Shixiang Wu, Jinjin Gu, Yihao Liu, Chao Dong, Yu~Qiao, and
  Chen Change~Loy.
\newblock Esrgan: Enhanced super-resolution generative adversarial networks.
\newblock In \emph{Proceedings of the European conference on computer vision
  (ECCV) workshops}, 2018.

\bibitem[Marriott et~al.(2021)Marriott, Romdhani, and Chen]{marriott20213d}
Richard~T Marriott, Sami Romdhani, and Liming Chen.
\newblock A 3d gan for improved large-pose facial recognition.
\newblock In \emph{CVPR}, 2021.

\bibitem[Tran et~al.(2017)Tran, Yin, and Liu]{tran2017disentangled}
Luan Tran, Xi~Yin, and Xiaoming Liu.
\newblock Disentangled representation learning gan for pose-invariant face
  recognition.
\newblock In \emph{CVPR}, 2017.

\bibitem[Pfau and Vinyals(2016)]{pfau2016connecting}
David Pfau and Oriol Vinyals.
\newblock Connecting generative adversarial networks and actor-critic methods.
\newblock \emph{arXiv:1610.01945}, 2016.

\bibitem[Chen et~al.(2020)Chen, Wang, Liu, Yang, Li, Wang, and
  Zhao]{chen2020computation}
Minshuo Chen, Yizhou Wang, Tianyi Liu, Zhuoran Yang, Xingguo Li, Zhaoran Wang,
  and Tuo Zhao.
\newblock On computation and generalization of generative adversarial imitation
  learning.
\newblock \emph{arXiv:2001.02792}, 2020.

\bibitem[Arjovsky et~al.(2017)Arjovsky, Chintala, and
  Bottou]{arjovsky2017wasserstein}
Martin Arjovsky, Soumith Chintala, and L{\'e}on Bottou.
\newblock Wasserstein generative adversarial networks.
\newblock In \emph{ICML}, pages 214--223, 2017.

\bibitem[Mao et~al.(2017)Mao, Li, Xie, Lau, Wang, and
  Paul~Smolley]{mao2017least}
Xudong Mao, Qing Li, Haoran Xie, Raymond~YK Lau, Zhen Wang, and Stephen
  Paul~Smolley.
\newblock Least squares generative adversarial networks.
\newblock In \emph{ICCV}, 2017.

\bibitem[Miyato et~al.(2018)Miyato, Kataoka, Koyama, and
  Yoshida]{miyato2018spectral}
Takeru Miyato, Toshiki Kataoka, Masanori Koyama, and Yuichi Yoshida.
\newblock Spectral normalization for generative adversarial networks.
\newblock \emph{arXiv preprint arXiv:1802.05957}, 2018.

\bibitem[Che et~al.(2016)Che, Li, Jacob, Bengio, and Li]{che2016mode}
Tong Che, Yanran Li, Athul~Paul Jacob, Yoshua Bengio, and Wenjie Li.
\newblock Mode regularized generative adversarial networks.
\newblock \emph{arXiv:1612.02136}, 2016.

\bibitem[Heusel et~al.(2017)Heusel, Ramsauer, Unterthiner, Nessler, and
  Hochreiter]{heusel2017gans}
Martin Heusel, Hubert Ramsauer, Thomas Unterthiner, Bernhard Nessler, and Sepp
  Hochreiter.
\newblock Gans trained by a two time-scale update rule converge to a local nash
  equilibrium.
\newblock \emph{Advances in neural information processing systems}, 30, 2017.

\bibitem[Metz et~al.(2016)Metz, Poole, Pfau, and
  Sohl-Dickstein]{metz2016unrolled}
Luke Metz, Ben Poole, David Pfau, and Jascha Sohl-Dickstein.
\newblock Unrolled generative adversarial networks.
\newblock \emph{arXiv:1611.02163}, 2016.

\bibitem[Farnia and Ozdaglar(2020)]{farnia2020gans}
Farzan Farnia and Asuman Ozdaglar.
\newblock Do gans always have nash equilibria?
\newblock In \emph{ICML}, 2020.

\bibitem[Salimans et~al.(2016)Salimans, Goodfellow, Zaremba, Cheung, Radford,
  and Chen]{salimans2016improved}
Tim Salimans, Ian Goodfellow, Wojciech Zaremba, Vicki Cheung, Alec Radford, and
  Xi~Chen.
\newblock Improved techniques for training gans.
\newblock \emph{NeurIPS}, 2016.

\bibitem[Tolstikhin et~al.(2017)Tolstikhin, Gelly, Bousquet, Simon-Gabriel, and
  Sch{\"o}lkopf]{tolstikhin2017adagan}
Ilya Tolstikhin, Sylvain Gelly, Olivier Bousquet, Carl-Johann Simon-Gabriel,
  and Bernhard Sch{\"o}lkopf.
\newblock Adagan: Boosting generative models.
\newblock \emph{arXiv:1701.02386}, 2017.

\bibitem[Pavan~Kumar and Jayagopal(2021)]{pavan2021generative}
MR~Pavan~Kumar and Prabhu Jayagopal.
\newblock Generative adversarial networks: a survey on applications and
  challenges.
\newblock \emph{International Journal of Multimedia Information Retrieval},
  10\penalty0 (1):\penalty0 1--24, 2021.

\bibitem[Daskalakis et~al.(2021)Daskalakis, Skoulakis, and
  Zampetakis]{daskalakis2021complexity}
Constantinos Daskalakis, Stratis Skoulakis, and Manolis Zampetakis.
\newblock The complexity of constrained min-max optimization.
\newblock In \emph{Proceedings of the 53rd Annual ACM SIGACT Symposium on
  Theory of Computing}, pages 1466--1478, 2021.

\bibitem[Mescheder et~al.(2018)Mescheder, Geiger, and
  Nowozin]{mescheder2018training}
Lars Mescheder, Andreas Geiger, and Sebastian Nowozin.
\newblock Which training methods for gans do actually converge?
\newblock In \emph{ICML}, 2018.

\bibitem[Gemp and Mahadevan(2018)]{gemp2018global}
Ian Gemp and Sridhar Mahadevan.
\newblock Global convergence to the equilibrium of gans using variational
  inequalities.
\newblock \emph{arXiv preprint arXiv:1808.01531}, 2018.

\bibitem[Engel et~al.(2017)Engel, Hoffman, and Roberts]{engel2017latent}
Jesse Engel, Matthew Hoffman, and Adam Roberts.
\newblock Latent constraints: Learning to generate conditionally from
  unconditional generative models.
\newblock \emph{arXiv preprint arXiv:1711.05772}, 2017.

\bibitem[Pedregosa(2016)]{pedregosa2016hyperparameter}
Fabian Pedregosa.
\newblock Hyperparameter optimization with approximate gradient.
\newblock In \emph{International conference on machine learning}, pages
  737--746. PMLR, 2016.

\bibitem[Lorraine et~al.(2020)Lorraine, Vicol, and
  Duvenaud]{lorraine2020optimizing}
Jonathan Lorraine, Paul Vicol, and David Duvenaud.
\newblock Optimizing millions of hyperparameters by implicit differentiation.
\newblock In \emph{International Conference on Artificial Intelligence and
  Statistics}, pages 1540--1552. PMLR, 2020.

\bibitem[Franceschi et~al.(2017)Franceschi, Donini, Frasconi, and
  Pontil]{franceschi2017forward}
Luca Franceschi, Michele Donini, Paolo Frasconi, and Massimiliano Pontil.
\newblock Forward and reverse gradient-based hyperparameter optimization.
\newblock In \emph{International Conference on Machine Learning}, pages
  1165--1173. PMLR, 2017.

\bibitem[Liu et~al.(2022)Liu, Mu, Yuan, Zeng, and Zhang]{liu2022general}
Risheng Liu, Pan Mu, Xiaoming Yuan, Shangzhi Zeng, and Jin Zhang.
\newblock A general descent aggregation framework for gradient-based bi-level
  optimization.
\newblock \emph{IEEE Transactions on Pattern Analysis and Machine
  Intelligence}, 2022.

\bibitem[Krizhevsky and Hinton(2010)]{krizhevsky2010convolutional}
Alex Krizhevsky and Geoff Hinton.
\newblock Convolutional deep belief networks on cifar-10.
\newblock \emph{Unpublished manuscript}, 40\penalty0 (7):\penalty0 1--9, 2010.

\bibitem[Krizhevsky et~al.(2009)Krizhevsky, Hinton,
  et~al.]{krizhevsky2009learning}
Alex Krizhevsky, Geoffrey Hinton, et~al.
\newblock Learning multiple layers of features from tiny images.
\newblock 2009.

\bibitem[Minguk~Kang and Park(2021)]{kang2021ReACGAN}
Minsu~Cho Minguk~Kang, Woohyeon~Shim and Jaesik Park.
\newblock {Rebooting ACGAN: Auxiliary Classifier GANs with Stable Training}.
\newblock 2021.

\bibitem[Gao et~al.(2018)Gao, Yang, Wang, Sun, Yang, and Zhou]{gao2018deep}
Fei Gao, Yue Yang, Jun Wang, Jinping Sun, Erfu Yang, and Huiyu Zhou.
\newblock A deep convolutional generative adversarial networks (dcgans)-based
  semi-supervised method for object recognition in synthetic aperture radar
  (sar) images.
\newblock \emph{Remote Sensing}, 10\penalty0 (6):\penalty0 846, 2018.

\end{thebibliography}

\end{document}